\documentclass{article}

\usepackage[main, final]{neurips_2026}

\usepackage{adjustbox}
\usepackage[utf8]{inputenc} 
\usepackage[T1]{fontenc}    
\usepackage[hidelinks]{hyperref}
\usepackage{xurl}            
\usepackage{booktabs}       
\usepackage{enumitem}
\usepackage{amsfonts, amsmath}       
\usepackage{nicefrac}       
\usepackage{microtype}      
\usepackage{graphicx}
\usepackage{xcolor}         
\usepackage[table]{xcolor}
\usepackage{multirow}
\usepackage{placeins}
\usepackage{siunitx}
\usepackage{tabularx}
\usepackage{subcaption}
\usepackage{tikz}
\hypersetup{
    colorlinks=true,
    linkcolor=black,
    citecolor=black,
    urlcolor=blue
}
\newcommand{\envsym}{\tikz[baseline=-0.15ex]{%
  \draw[line width=0.35pt] (0,0) rectangle (0.75em,0.5em);
  \draw[line width=0.35pt] (0,0.5em) -- (0.375em,0.2em) -- (0.75em,0.5em);
}}
\title{Beyond LoRA vs. Full Fine-Tuning: Gradient-Guided Optimizer Routing for LLM Adaptation}

\author{
Haozhan Tang$^{1,2,\envsym}$ \quad
Xiuqi Zhu$^{1,*}$ \quad
Xinyin Zhang$^{1,*}$ \\[3pt]
\bfseries Boxun Li$^{3}$ \quad
Virginia Smith$^{1}$ \quad
Kevin Kuo$^{1,\envsym}$\\[5pt]
\normalfont $^{1}$Carnegie Mellon University \quad $^{2}$Tsinghua University \quad $^{3}$Infinigence AI \\
}

\begin{document}

\maketitle

\begingroup
\renewcommand{\thefootnote}{}
\footnotetext[0]{$^{*}$These authors contributed equally.}
\footnotetext[0]{$^{\envsym}$Corresponding authors: \texttt{hztang801@gmail.com}, \texttt{kkuo2@andrew.cmu.edu}.}
\endgroup

\begin{abstract}
Recent literature on fine-tuning Large Language Models highlights a fundamental debate. While Full Fine-Tuning (FFT) provides greater representational plasticity, Low-Rank Adaptation (LoRA) provides regularizing effects and can potentially surpass the performance of FFT.
Through empirical evaluation across diverse tasks (SQL, Medical QA, and Counterfactual Knowledge) and varying language models (Gemma-3-1B, Qwen2.5-1.5B, and Qwen2.5-3B), we 
find that the better static architecture depends on the task and model. 
To address this challenge, we propose a Mixture of LoRA and Full (MoLF) Fine-Tuning, a unified framework that enables continuous navigation between both training regimes. MoLF dynamically routes updates between FFT and LoRA at the optimizer level to ensure that exact gradient signals are available to both experts throughout training, while only selected experts update their weights. For memory-constrained environments, we also introduce MoLF-Efficient, which freezes base weights and only routes updates among a pair of LoRA experts of potentially varying rank. Our evaluations show that MoLF either improves on or stays within $1.5$ percentage points of the better of FFT and LoRA across the nine tested settings, while MoLF-Efficient outperforms both AdaLoRA and AdaMix in eight of nine settings, with gains over the stronger baseline of up to $11.70$ percentage points on Fact, $3.13$ on Med, and $2.98$ on SQL.
Our code is open-sourced at \url{https://github.com/11785T23/molf.git}.
\end{abstract}

\section{Introduction}





Fine-tuning is an effective approach to improve pre-trained Large Language Models (LLMs) on downstream NLP tasks~\citep{brown2020language,touvron2023llama,touvron2023llama2,chung2024scaling}. However, LLMs are typically overparameterized relative to the quantity of available training data, which creates a tension between adaptation and generalization: aggressive optimization can overfit or degrade pretrained representations~\citep{jiang2020smart,aghajanyan2020better}. A natural way to control this trade-off is through varying the fine-tuning parameterization. In this work, we study two standard fine-tuning methods: Full Fine-Tuning (FFT) and Low-Rank Adaptation (LoRA)~\citep{hu2022lora}.


\begin{figure*}[ht]
    \centering
    \includegraphics[width=\textwidth]{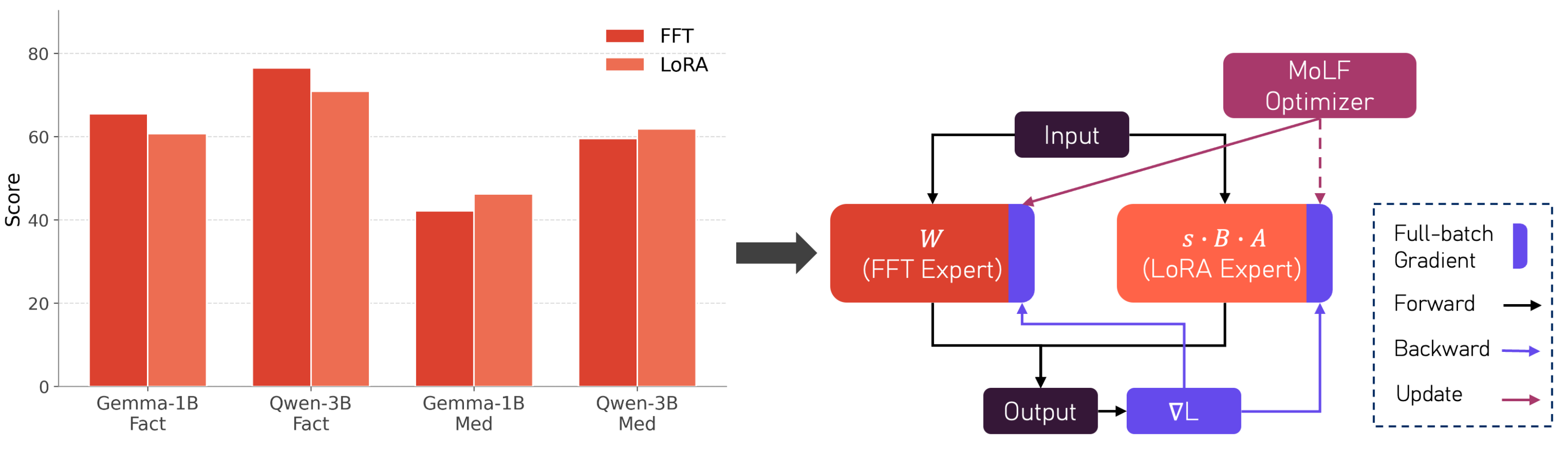}
    \caption{Our evaluations across the tested settings show task-dependent FFT/LoRA preferences: FFT scores higher on Fact, tuned LoRA scores higher on Med, and both perform similarly on SQL. Capacity and regularization provide possible explanations, rather than established mechanisms; the endpoint analysis in Appendix~\ref{app:spectral} tests their limits. The Mixture of LoRA and Full (MoLF) framework routes physical parameter updates between dense and low-rank experts at the optimizer level. Every expert participates in forward and backward computation and accumulates AdamW moments from the full batch.\looseness=-1}
    \label{fig:overview}
\end{figure*}

FFT's greater capacity motivates LoRA extensions that mix same-rank modules~\citep{wang2022adamix,albert2025randlora} or adapt ranks during training~\citep{zhang2023adaptive,zhang2023increlora,liu2024alora}. Yet low-rank constraints can regularize adaptation, allowing LoRA to outperform FFT~\citep{hu2022lora,biderman2024lora}. 
Taken together, these prior works motivate an architecture which selectively leverages both dense and low-rank updates to maximize adaptation quality.

We propose a Mixture of LoRA and Full fine-tuning (MoLF), which treats FFT and LoRA as candidate update \emph{experts}. Whereas mixture-of-PEFT and adaptive LoRA methods compute updates within low-rank modules~\citep{wang2022adamix,zhang2023adaptive}, MoLF selects between dense and low-rank updates locally. Both experts participate in every forward/backward pass and accumulate AdamW moments; a gradient-informed score then selects one physical update per module. The parameterization and optimizer-state layout remain fixed, allowing an unselected expert to resume from current moments. This update sparsity retains both experts' computation \mbox{and optimizer-state costs.}

For memory-constrained settings, MoLF-Efficient (MoLF-E) freezes the dense model and routes updates among two LoRA experts. It applies the same optimizer-level selection within a LoRA-only parameterization, trading full-rank expressiveness for lower memory use. Our contributions are:
\begin{enumerate}
    \item We extensively tune FFT and LoRA in 9 settings where we fine-tune 3 LLMs (Gemma-3-1B, Qwen2.5-1.5B, Qwen2.5-3B) on 3 datasets (CounterFact, MedMCQA, and Text-to-SQL). The best-performing method and rank vary across the tested settings, suggesting that methods should not simply seek to maximize the effective rank of the architecture. These point estimates motivate the comparison; they do not establish a universal task taxonomy.
    \item We propose MoLF, which unifies FFT and LoRA through optimizer-level update routing. MoLF fine-tunes both experts while constraining physical updates using a momentum-based, parameter-normalized scoring heuristic. Across the 9 tested settings, its scores are above or within $1.5$ percentage points of the better tuned baseline (FFT or LoRA); this is a statement about the reported point estimates, rather than \mbox{a guarantee across training seeds.}
    \item We propose MoLF-E, a memory-efficient variant which freezes the base model and routes updates among a pair of LoRA experts. With two rank-$128$ experts, MoLF-E scores higher than both AdaLoRA and AdaMix in eight of nine settings, with a maximum margin of $20.01$ percentage points over AdaMix on Fact.
\end{enumerate}

A Meta-Llama-3-8B mathematical-reasoning experiment extends the evaluation beyond the original models and tasks (Section~\ref{subsec:math_results}). Additional studies examine training-seed variability (Appendix~\ref{app:seed_study}) and workload-dependent time and memory costs at 7B/8B scale (Appendix~\ref{app:efficiency}).

\section{Related Work}

\textbf{FFT versus LoRA.} Prior work provides evidence that some pre-trained LLM fine-tuning tasks admit effective adaptation in low-dimensional subspaces, helping explain why LoRA can be effective~\citep{hu2022lora,aghajanyan2021intrinsic,schulman2025lora}. Follow-up works have tried to further improve LoRA by mimicking FFT or by increasing the effective rank of LoRA~\citep{albert2025randlora,hao2024flora,wang2024lora,lialin2024relora}. Empirical comparisons between the two methods yield mixed conclusions: some work finds that LoRA matches or exceeds FFT, with the low-rank constraint acting as an implicit regularizer that mitigates forgetting and reduces reliance on explicit KL penalties during RLHF~\citep{hu2022lora,biderman2024lora,sun2023exploring,du2024study}. Conversely, other work finds that FFT outperforms LoRA, particularly in instruction tuning and knowledge-intensive settings~\citep{ivison2023camels,pletenev2025much}. Beyond raw accuracy, FFT and LoRA also differ in the structure of their learned solutions and their robustness to distribution shift~\citep{biderman2024lora,shuttleworth2025lora}. These observations motivate measuring task-dependent preferences without assuming that one mechanism explains every setting.

\textbf{Adaptive LoRA.} LoRA is a parameter-efficient fine-tuning (PEFT) method which injects trainable low-rank matrices into a frozen base model~\citep{hu2022lora}. Despite its efficiency, LoRA is sensitive to the choice of rank, motivating a line of work on methods that use importance scores (e.g. parameter or gradient norms) to promote or prune rank components dynamically across layers~\citep{zhang2023adaptive,zhang2023increlora,liu2024alora,chang2025elalora}. A similar family of methods takes a finer-grained approach by decomposing LoRA updates into rank-1 components and selectively gating or routing over them, either via sparse regularization, meta-learning, or importance-based pruning~\citep{ding2023sparse,zhang2024autolora,mao2024dora}. Finally, a related line of work aims to produce LoRA modules that are robust to rank truncation at inference time~\citep{valipour2023dylora,rajabzadeh2024qdylora}.

\textbf{Mixture-of-PEFT.} Mixture-of-Experts (MoE) models maintain multiple parallel sub-networks (experts) and route each input to a subset of them~\citep{jacobs1991adaptive,shazeer2017outrageously}. Several works combine PEFT with MoE-style routing, treating each LoRA adapter as an expert. These works are motivated by two related but distinct goals. First, a pool of adapters can increase representational capacity while controlling how many adapters are evaluated per input~\citep{wang2022adamix,zhu2023sira,adamole2024}. Second, when specializing to multiple domains, a shared LoRA can suffer from gradient conflicts and negative transfer, motivating domain- or task-specific experts~\citep{zadouri2024molora,wu2024mixture,li2024mixlora,dou2024loramoe}. MoLF adds a dense update candidate to this low-rank choice and selects after backward computation. MoLF-E remains LoRA-only, with continuous gradient observation and sparse physical updates.

\textbf{Coordinate-sparse fine-tuning.} Sparse fine-tuning updates only a selected subset of pre-trained parameters, constraining the support of the update rather than its rank~\citep{guo2021diffpruning,sung2021fish,ansell2022ltsft}. Recent work scales this approach to LLMs and reports performance competitive with LoRA~\citep{ansell2024spiel,yang2024s2ft}. A matched FFT/sparse-fine-tuning comparison remains future work; our experiments do not establish LoRA as the best constrained counterpart to FFT (Appendix~\ref{sec:limitations}).

\section{Understanding Fine-Tuning Dynamics: An Empirical Analysis}

\label{sec:fft_vs_lft}
Recent literature emphasizes different aspects of the fine-tuning dynamics of LLMs. One line~\citep{hu2022lora,aghajanyan2021intrinsic,schulman2025lora} studies how low-dimensional or low-rank adaptation can efficiently capture useful updates. A competing line~\citep{biderman2024lora} documents settings in which unconstrained Full Fine-Tuning (FFT) learns more than Low-Rank Adaptation (LoRA), while LoRA retains more pretrained capabilities: \textit{``LoRA Learns Less and Forgets Less.''} These findings motivate studying when the capacity and regularization effects of the two parameterizations are useful, without assuming that one is universally preferable.

We empirically evaluate FFT and LoRA across tasks and model scales. Our setup includes Google Gemma-3-1B, Qwen2.5-1.5B, and Qwen2.5-3B, evaluated on Factual Knowledge (CounterFact~\citep{meng2022locating}), Medical QA (MedMCQA~\citep{pal2022medmcqa}), and Text-to-SQL (Gretel synthetic Text-to-SQL~\citep{gretel2024synthsql}). These tasks differ in their objectives; we do not assume their intrinsic dimensionalities in advance. Following a hyperparameter sweep over learning rates, schedulers, and LoRA ranks ($r \in \{8, 16, 32, 64, 128\}$), we report the resulting scores for each tested setting in Table~\ref{tab:fft_vs_lft}. The sweep details are in Appendix~\ref{subsec:hyperparam_sweep}.

\begin{table}[ht]
\centering
\caption{Fine-Tuning Performance Benchmark. Cells are Efficacy Score (\%) on Fact and accuracy (\%) on Med and SQL, rounded to two decimal places. The best score in each row is in bold, and any score within a 1.5-percentage-point margin of the best is highlighted in blue.}
\label{tab:fft_vs_lft}
\resizebox{\textwidth}{!}{%
\begin{tabular}{@{}llcccccc@{}}
\toprule
\textbf{Benchmark} & \textbf{Model} & \textbf{FFT} & \textbf{LoRA ($r=8$)} & \textbf{LoRA ($r=16$)} & \textbf{LoRA ($r=32$)} & \textbf{LoRA ($r=64$)} & \textbf{LoRA ($r=128$)} \\ \midrule
\multirow{3}{*}{\textbf{Fact}} & \text{Gemma-1B} & \textbf{65.50} & 60.05 & 60.77 & 60.49 & 54.43 & 54.63 \\
 & \text{Qwen-1.5B} & \textbf{74.95} & 68.39 & 69.05 & 65.89 & 65.34 & 70.78 \\
 & \text{Qwen-3B} & \textbf{76.54} & 70.56 & 66.88 & 61.32 & 65.08 & 70.85 \\ \midrule
\multirow{3}{*}{\textbf{Med}} & \text{Gemma-1B} & 42.15 & 40.90 & 42.10 & 43.65 & \textcolor{blue}{46.09} & \textbf{46.19} \\
 & \text{Qwen-1.5B} & 55.39 & 54.03 & 54.48 & 54.46 & 55.99 & \textbf{57.66} \\
 & \text{Qwen-3B} & 59.55 & 58.95 & 59.96 & \textcolor{blue}{60.67} & \textcolor{blue}{60.84} & \textbf{61.85} \\ \midrule
\multirow{3}{*}{\textbf{SQL}} & \text{Gemma-1B} & \textbf{72.76} & \textcolor{blue}{71.82} & \textcolor{blue}{71.93} & \textcolor{blue}{71.77} & \textcolor{blue}{71.68} & \textcolor{blue}{71.26} \\
 & \text{Qwen-1.5B} & \textbf{74.81} & \textcolor{blue}{73.95} & \textcolor{blue}{74.20} & \textcolor{blue}{74.39} & \textcolor{blue}{74.02} & \textcolor{blue}{73.98} \\
 & \text{Qwen-3B} & \textcolor{blue}{75.06} & \textbf{75.23} & \textcolor{blue}{74.97} & \textcolor{blue}{74.97} & \textcolor{blue}{74.17} & \textcolor{blue}{74.42} \\ \bottomrule
\end{tabular}%
}
\end{table}

The results exhibit three patterns. Capacity and regularization offer possible explanations, with their mathematical intuition and caveats given in Appendix~\ref{app:rank_mechanisms}; the task scores alone \mbox{do not establish these mechanisms.}

\begin{itemize}
    \item \textbf{Fact (FFT advantage and a possible capacity/optimization constraint):}\\
    FFT exceeds every swept LoRA rank by at least 4.17 percentage points across all models. Rank-approximation bounds~\citep{eckart1936approximation,mirsky1960symmetric} concern parameter error, not task loss; the endpoint evidence below contradicts a heavier-tail explanation for Fact. Low-rank constraints may nevertheless discard task-sensitive directions or alter the optimization trajectory: training LoRA differs from compressing a completed FFT run.

    \item \textbf{Med (LoRA advantage and a regularization hypothesis):}\\
    Higher-rank LoRA outperforms FFT in the tested settings, consistent with a possible regularization benefit. Its factor parameterization filters and rescales gradient components through evolving subspaces, potentially attenuating noise. This intuition does not establish that retained components are useful signal, and AdamW's adaptive preconditioning further modifies the updates. Because we do not directly measure pretrained-capability retention, reduced forgetting remains a hypothesis rather than an established \mbox{explanation for the Med scores.}

    \item \textbf{SQL (Empirical rank insensitivity):}\\
    FFT and LoRA achieve similar accuracy, with little variation across the swept ranks. This empirical rank insensitivity does not establish low intrinsic dimension: the measured FFT endpoint retains substantial energy beyond rank 128, yet low-rank adaptation achieves comparable task performance.
\end{itemize}

\paragraph{Direct endpoint and task-score evidence.}
We analyze the archived, sweep-selected Qwen2.5-1.5B FFT endpoints, covering seven projection families in all 28 layers (196 matrices per task). For each displacement $\Delta W=W^{\mathrm{FFT}}-W_{\mathrm{base}}$, let $T_{128}$ be the fraction of squared Frobenius energy outside its best rank-128 SVD approximation. We also evaluate checkpoints in which the analyzed FFT displacements are replaced by their rank-128 approximations. Table~\ref{tab:spectral_main} reports the resulting task-score loss alongside the median parameter-space residual.

\begin{table}[ht]
\centering
\caption{Endpoint compressibility and task-score sensitivity on Qwen2.5-1.5B. $T_{128}$ is the median over 196 projection matrices. The absolute gap is the full FFT score minus the score after rank-128 endpoint truncation; the normalized gap divides by the full FFT gain over the pretrained model. These results use one FFT endpoint per task, not training-seed averages.}
\label{tab:spectral_main}
\begin{tabular}{@{}lccc@{}}
\toprule
Task & Median $T_{128}$ & Gap (points) & Gap / total FFT gain \\
\midrule
Fact & 0.450 & 4.92 & 8.5\% \\
Med  & 0.639 & 1.41 & 14.1\% \\
SQL  & 0.598 & 1.70 & 14.7\% \\
\bottomrule
\end{tabular}
\end{table}

Fact has the most Frobenius-compressible endpoint of the three tasks, but its rank-128 truncation incurs the largest \emph{absolute} score loss. The predeclared median paired Fact-minus-SQL $T_{128}$ contrast is $-0.1093$ (95\% layer-cluster bootstrap interval $[-0.1266,-0.0990]$), contrary to a heavier-tail explanation for Fact. Normalizing the score loss by each task's total FFT gain reverses the task ordering, and the interval describes layer heterogeneity for one selected endpoint pair, not training-seed uncertainty. Appendix~\ref{app:spectral} provides checkpoint provenance, definitions, additional rank and subspace diagnostics, the complete truncation scores, and the bootstrap protocol.

The revised interpretation is that endpoint Frobenius energy alone cannot explain the task-dependent performance differences. Sensitivity to discarded directions and the optimization constraints of a low-rank trajectory are plausible complementary explanations, which these measurements do not causally separate. Together, the benchmark and diagnostics motivate adapting the update parameterization across modules and tasks rather than assuming one static choice is universally preferable.

\section{Methodology: Mixture of LoRA and Full Fine-Tuning}
Mixture of LoRA and Full (MoLF) Fine-Tuning routes physical parameter updates between dense and low-rank experts using their optimizer moments. We also introduce its memory-constrained variant, MoLF-Efficient. Here, an \emph{expert} denotes a candidate update expert rather than a \mbox{token-routed subnetwork.}

\subsection{Background: LoRA and Mixture-of-Experts}
Low-Rank Adaptation (LoRA)~\citep{hu2022lora} reduces fine-tuning memory by freezing pre-trained weights $W_{\text{base}} \in \mathbb{R}^{d_{\text{out}} \times d_{\text{in}}}$ and training low-rank matrices $A \in \mathbb{R}^{r \times d_{\text{in}}}, B \in \mathbb{R}^{d_{\text{out}} \times r}$. The bottleneck $r$ bounds the rank of $BA$ and constrains optimization; its effect on task performance cannot be determined from endpoint rank alone (Appendix~\ref{app:spectral}).

\begin{figure}[htbp]
\centering
\includegraphics[width=0.9\textwidth]{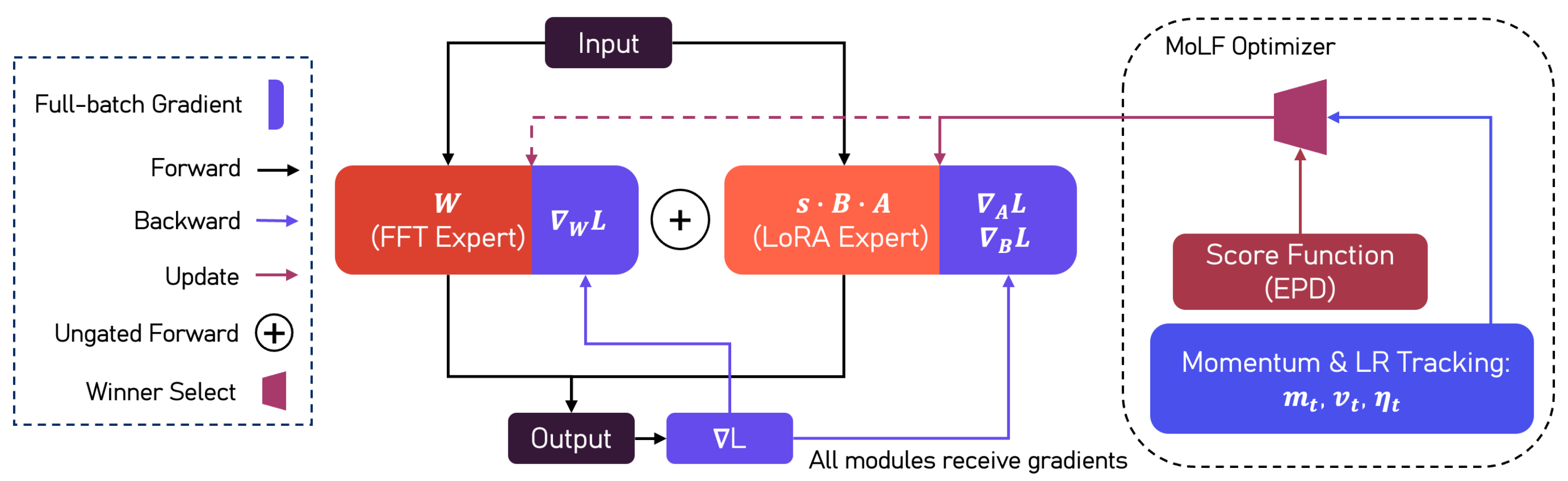}
\caption{Overview of MoLF. Both candidate experts participate in forward and backward computation and accumulate AdamW moments at every step. The module-local Top-$K$ selector acts only on physical parameter updates; an unselected expert retains its weights. Standard MoLF uses one FFT and one LoRA candidate with $K=1$.}
\label{fig:methodology}
\end{figure}

Mixture-of-Experts (MoE) architectures~\citep{jacobs1991adaptive,shazeer2017outrageously} conditionally route tokens through independent experts, making data exposure and load balancing depend on the routing policy. MoLF (Figure~\ref{fig:methodology}) places selection after backward computation: all experts process the batch, and only physical updates are sparse. Its routing therefore provides no conditional-computation savings.

\FloatBarrier

\subsection{MoLF Architecture and Inference}
MoLF represents each linear projection as an unconditional sum of experts. For input $x$, the forward pass evaluates:
\begin{equation}
y = W_{\text{base}} x + \sum_{i=1}^{N} \frac{\alpha_i}{\sqrt{r_i}} B_i \big(A_i (\text{Dropout}(x))\big).
\label{eq:forward_pass}
\end{equation}
The dense matrix $W_{\text{base}}$ is the FFT expert; each independently parameterized pair $(A_i, B_i)$ is a rank-$r_i$ LoRA expert with Rank-Stabilized LoRA (RS-LoRA) scaling $\alpha_i / \sqrt{r_i}$~\citep{kalajdzievski2023rank}. Standard MoLF uses $N=1$: one dense and one rank-$128$ LoRA candidate per wrapped module. All gradients come from this joint model, not separately trained FFT or LoRA models.

\subsection{Dynamic Gradient Routing via Sparse AdamW}
Our sparse AdamW optimizer~\citep{kingma2014adam, loshchilov2017decoupled} separates moment tracking from weight modification \mbox{within each wrapped module.}

\subsection*{Phase 1: Universal Momentum Tracking}
Let $i=0$ denote the FFT expert $W_{\text{base}}$ and $i \in \{1,\dots,N\}$ the LoRA experts within a module. At every step $t$, all experts receive batch-averaged gradients $g_t^{(i)}$. The optimizer updates the exponential moving averages of their first moments $m_t^{(i)}$ and uncentered second moments $v_t^{(i)}$:
\begin{align}
m_t^{(i)} &= \beta_1 m_{t-1}^{(i)} + (1 - \beta_1) g_t^{(i)} \label{eq:first_moment} \\
v_t^{(i)} &= \beta_2 v_{t-1}^{(i)} + (1 - \beta_2) \left(g_t^{(i)}\right)^2 \label{eq:second_moment}
\end{align}

All Adam step counters advance regardless of selection, keeping bias corrections ($1-\beta_1^t, 1-\beta_2^t$) synchronized. An expert whose weights remain unchanged retains current gradient statistics, rather than starting with newly initialized moments when it next receives an update. This design alone does not guarantee lower training variance than other routing or rank-adaptation methods.

\subsection*{Phase 2: Expert Scoring by Expected Preconditioned Descent}
\label{subsec:epd_score}
Our scoring heuristic approximates the loss reduction associated with a candidate's preconditioned update. Because AdamW does not descend along $g_t^{(i)}$ but along the direction determined by $m_t^{(i)} / (\sqrt{v_t^{(i)}} + \epsilon)$ and bias correction, raw gradient norms omit the per-coordinate adaptive scaling that the optimizer uses.

The Expected Preconditioned Descent (EPD) score $\mathcal{S}_t^{(i)}$ is a first-order surrogate normalized by the candidate's parameter count $N_{\text{params}}^{(i)}$. Replacing the current gradient by its first moment gives:
\begin{equation}
\mathcal{S}_t^{(i)} = \frac{\eta_t^{(i)}}{N_{\text{params}}^{(i)}} \sum_{\theta \in \Theta_i} \frac{\left(m_t^{(i)}\right)^2}{\sqrt{v_t^{(i)}} + \epsilon}
\label{eq:epd_score}
\end{equation}
Here $\eta_t^{(i)}$ is the actual scheduled learning rate of expert $i$:
\begin{equation}
\eta_t^{(i)} = c_t\,\eta_{\mathrm{peak}}^{(i)}.
\label{eq:scheduled_lr}
\end{equation}
The multiplier $c_t$ follows a shared linear or cosine schedule; each peak rate $\eta_{\mathrm{peak}}^{(i)}$ is configured rather than learned. The original sweep uses a $0.1$ warmup ratio. The Taylor motivation (Appendix~\ref{sec:epd_derivation}) approximates $g_t$ by $m_t$ and omits higher-order terms and decoupled weight decay. EPD is a nonnegative heuristic, not an unbiased estimate or a guarantee of the largest realized loss reduction.

The original nine-setting evaluation fixed LoRA's peak rate at $5\times10^{-5}$ and selected the dense peak rate from $\{10^{-4},5\times10^{-5},2\times10^{-5}\}$ jointly with the scheduler (Appendix~\ref{subsec:hyperparam_sweep}). Sharing a conventional learning rate across experts removes explicit relative-rate preference while leaving rank configurable. Both MoLF experts use $5\times10^{-5}$ in the five-seed CounterFact and 8B mathematical-reasoning experiments (Appendix~\ref{app:seed_study} and Section~\ref{subsec:math_results}). These results do not show that this default replaces configuration search in all nine original settings.

EPD has two scaling properties (Equation~\ref{eq:epd_score}). First, it is not rank-invariant. Under a uniform-rescaling approximation, RS-LoRA scaling $\alpha_i / \sqrt{r_i}$ scales per-coordinate gradients and $m_t^2/\sqrt{v_t}$ proportionally. Summing over $N_{\text{params}}^{(i)}\propto r_i$ coordinates and normalizing by this count yields approximate score scaling $\eta_t^{(i)}\alpha_i/\sqrt{r_i}$. At equal learning rates and scaling constants, a rank-$r$ candidate can therefore receive a $\sqrt{128/r}$ score advantage over a rank-$128$ candidate. This approximation excludes changes in learned factors or gradient distributions across ranks. Candidate size therefore affects the score without implying an optimal rank preference. Second, retaining the actual learning rate allows identical moment statistics to yield different selections \mbox{when expert rates differ.}

The Preconditioned Frobenius Norm (PFN; Appendix~\ref{app:pfn_details}) is an author-constructed diagnostic control using the RMS preconditioned Adam direction. Uniform gradient rescaling approximately cancels in its moment ratio; the learning-rate multiplier is omitted. PFN thus tests a simpler control against EPD's descent-magnitude information without isolating the learning-rate term alone. The fixed-rank ablation does not test unequal-rank routing, and the MoLF-E rank sweep entangles capacity with routing preference. Rank-invariant scores, rank-aware rates, and routing priors remain untested (Appendices~\ref{subsec:ablation_rank} and~\ref{sec:limitations}).

\subsection*{Phase 3: Top-K Sparse AdamW Update}
Within each module, the Top-$K$ candidates by EPD score receive the AdamW update~\citep{loshchilov2017decoupled}, using their tracked moments, scheduled rates $\eta_t^{(i)}$, and decoupled weight decay $\lambda_i$:
\begin{equation}
\theta_{t}^{(i)} \leftarrow \theta_{t-1}^{(i)} \left(1 - \eta_t^{(i)} \lambda_i \right) - \frac{\eta_t^{(i)}}{1-\beta_1^t} \left( \frac{m_t^{(i)}}{\sqrt{v_t^{(i)} / (1-\beta_2^t)} + \epsilon} \right) \quad \text{for } i \in \text{Winners}.
\label{eq:topk_update}
\end{equation}

Unselected candidates retain their weights ($\theta_t^{(i)} = \theta_{t-1}^{(i)}$), including skipping decay. Each projection routes independently, allowing different choices within a transformer layer.

For standard MoLF's two candidates, $K=1$ is the only nontrivial sparse choice; $K=2$ updates both in the ``no update routing'' ablation. This comparison changes both selection and physical update count, while retaining all candidates' forward/backward computation, gradients, and moment tracking. Non-routed parameters follow ordinary updates in either setting.

Larger candidate pools add forward/backward work, gradients, and optimizer state. Their benefits, choice of $K$, rank calibration, and update balancing remain empirical questions. Appendix~\ref{app:efficiency} reports costs for the studied pair.

\subsection{MoLF-Efficient: Adaptive LoRA-Only Mixtures}

Standard MoLF tracks dense gradients and Adam states for $W_{\text{base}}$. For memory-constrained hardware, MoLF-Efficient (MoLF-E) freezes $W_{\text{base}}$ and all non-linear parameters and excludes them from optimizer state tracking. The reported configuration uses two independently initialized rank-$128$ LoRA adapters with Top-$1$ routing; Appendix~\ref{subsec:ablation_rank} varies the first adapter's rank.

Without a trainable FFT expert, the same EPD heuristic selects between low-rank candidates. Independent initialization permits different optimization trajectories under the moment-tracking rule above. MoLF-E therefore tests optimizer-level routing with reduced dense-state memory, but does not choose between FFT and LoRA or retain standard MoLF's full-rank expressiveness. Its memory benefit does not imply faster training (Appendix~\ref{app:efficiency}).

\subsection{Post-Training Fusion and Zero-Overhead Inference}

After fine-tuning, all LoRA experts merge algebraically into their dense base weights:
\begin{equation}
W_{\text{final}} = W_{\text{base}} + \sum_{i=1}^N \frac{\alpha_i}{\sqrt{r_i}} \left(B_i A_i\right). 
\label{eq:fuse}
\end{equation}

With dropout disabled, the merged model has the base LLM's structure and uses standard inference engines without separate adapters or routing. This inference-time property does not reduce the training costs of maintaining both experts.

\section{Results}
\label{sec:results}
 
\subsection{Experimental Setup}
\label{subsec:setup}
Our original evaluation covers CounterFact, MedMCQA, and Gretel Text-to-SQL (Section~\ref{sec:fft_vs_lft}) on Gemma-3-1B~\citep{team2025gemma}, Qwen2.5-1.5B, and Qwen2.5-3B~\citep{yang2024qwen2_5}. We compare MoLF and MoLF-E with FFT, LoRA at ranks $r\in\{8,16,32,64,128\}$, AdaLoRA~\citep{zhang2023adaptive}, and AdaMix~\citep{wang2022adamix}, using AdamW~\citep{loshchilov2017decoupled}. Appendix~\ref{subsec:hyperparam_sweep} documents configuration selection. We report Efficacy Score (ES) on Fact and accuracy on Med/SQL, all as percentages (Appendix~\ref{subsec:metrics}). Each method's result in the original nine tested settings comes from one training run; additional seed, reasoning, and efficiency studies have separately specified protocols.
 
\subsection{MoLF vs.\ Full Fine-Tuning and Tuned LoRA}
\label{subsec:main_results}

\begin{table}[ht]
\centering
\large
\caption{MoLF vs.\ Full Fine-Tuning and LoRA (best rank). Cells are Efficacy Scores (\%) on Fact and accuracies (\%) on Med and SQL. The best score in each column is in bold; any score within a $1.5$ percentage-point margin of the best baseline (FFT or LoRA) is highlighted in blue. Checkmarks indicate point-estimate performance recovery within this margin, \mbox{not statistical equivalence.}}
\label{tab:molf_vs_static}
\adjustbox{max width=\textwidth}{%
\begin{tabular}{@{}lccccccccc@{}}
\toprule
\multirow{2}{*}{\textbf{Method}} & \multicolumn{3}{c}{\textbf{Fact}} & \multicolumn{3}{c}{\textbf{Med}} & \multicolumn{3}{c}{\textbf{SQL}} \\ \cmidrule(lr){2-4} \cmidrule(lr){5-7} \cmidrule(l){8-10}
 & \textbf{Gemma-1B} & \textbf{Qwen-1.5B} & \textbf{Qwen-3B} & \textbf{Gemma-1B} & \textbf{Qwen-1.5B} & \textbf{Qwen-3B} & \textbf{Gemma-1B} & \textbf{Qwen-1.5B} & \textbf{Qwen-3B} \\ \midrule
\textbf{Best of FFT/LoRA} & 65.50 & 74.95 & 76.54 & 46.19 & 57.66 & 61.85 & 72.76 & 74.81 & 75.23 \\ \midrule
\textbf{FFT} & \textbf{65.50} & \textcolor{blue}{74.95} & \textbf{76.54} & 42.15 & 55.39 & 59.55 & \textcolor{blue}{72.76} & \textbf{74.81} & \textcolor{blue}{75.06} \\
$\hookrightarrow$ within 1.5 points of best & \checkmark & \checkmark & \checkmark & & & & \checkmark & \checkmark & \checkmark \\ \midrule
\textbf{LoRA (best rank)} & 60.77 & 70.78 & 70.85 & \textbf{46.19} & \textbf{57.66} & \textbf{61.85} & \textcolor{blue}{71.93} & \textcolor{blue}{74.39} & \textcolor{blue}{75.23} \\
$\hookrightarrow$ within 1.5 points of best & & & & \checkmark & \checkmark & \checkmark & \checkmark & \checkmark & \checkmark \\ \midrule
\textbf{MoLF} & \textcolor{blue}{64.88} & \textbf{75.15} & \textcolor{blue}{76.15} & \textcolor{blue}{45.40} & \textcolor{blue}{56.28} & \textcolor{blue}{60.60} & \textbf{73.09} & \textcolor{blue}{74.44} & \textbf{75.41} \\
$\hookrightarrow$ within 1.5 points of best & \checkmark & \checkmark & \checkmark & \checkmark & \checkmark & \checkmark & \checkmark & \checkmark & \checkmark \\ \bottomrule
\end{tabular}%
}
\end{table}
 
Table~\ref{tab:molf_vs_static} shows that MoLF's score is above or within $1.5$ points of the better static baseline in all nine tested settings and is highest in three. FFT trails tuned LoRA by up to $4.04$ points on Med, while LoRA trails FFT by up to $5.69$ points on Fact. MoLF approaches FFT on Fact, exceeds it by up to $3.25$ points on Med, and has the highest SQL score on two models. These comparisons across the tested settings support module-local update selection; they do not establish \mbox{robustness across training seeds.}

Appendix~\ref{app:seed_study} reports five paired CounterFact seeds: MoLF has a higher observed mean than FFT, but their paired 95\% interval includes zero and they use different selected learning rates. The appendix gives the full results and LoRA control.

\subsection{Mathematical Reasoning at 8B Scale}
\label{subsec:math_results}
We fine-tune Meta-Llama-3-8B~\citep{grattafiori2024llama} for one epoch on 34,702 ODA-Mixture-100k examples~\citep{opendataarena2026mixture} that fit its native 8,192-token context. FFT, LoRA-128, and both MoLF experts use the same $5\times10^{-5}$ learning rate. Table~\ref{tab:math_8b_main} reports MATH-500 accuracy~\citep{hendrycks2021measuring,huggingfaceh4math500}, averaged over three generation seeds for one training seed per method, together with the fraction of generations reaching the 7,168-token output cap. The data filtering, packing, evaluation protocol, and bootstrap procedure are \mbox{detailed in Appendix~\ref{app:math_protocol}.}

\begin{table}[htbp]
\centering
\caption{Meta-Llama-3-8B mathematical reasoning after ODA fine-tuning at a shared learning rate. The three repeats are generation seeds, not independent training runs.}
\label{tab:math_8b_main}
\begin{tabular}{lcc}
\toprule
Method & MATH-500 accuracy (\%) & Cap hit (\%) \\
\midrule
FFT & 32.93 & 58.53 \\
MoLF & 31.53 & 60.07 \\
LoRA-128 & 20.60 & 70.20 \\
\bottomrule
\end{tabular}
\end{table}

MoLF exceeds matched LoRA by $10.93$ points (paired problem-bootstrap 95\% interval $[8.40,\,13.40]$) and trails FFT by $1.40$ points (MoLF minus FFT: $[-3.87,\,1.00]$). The latter interval includes zero but does not establish equivalence. These intervals describe evaluation-problem variability conditional on the trained checkpoints, not training-seed uncertainty. FFT and MoLF have similar cap-hit rates; LoRA reaches the cap more often, so termination behavior may contribute to the gap. This single 8B setting does not establish general superiority on reasoning or instruction-following tasks.
 
\subsection{MoLF-E vs.\ Adaptive PEFT Baselines}
\label{subsec:molf_efficient}
\begin{figure}[htbp]
\centering
\includegraphics[width=0.95\textwidth]{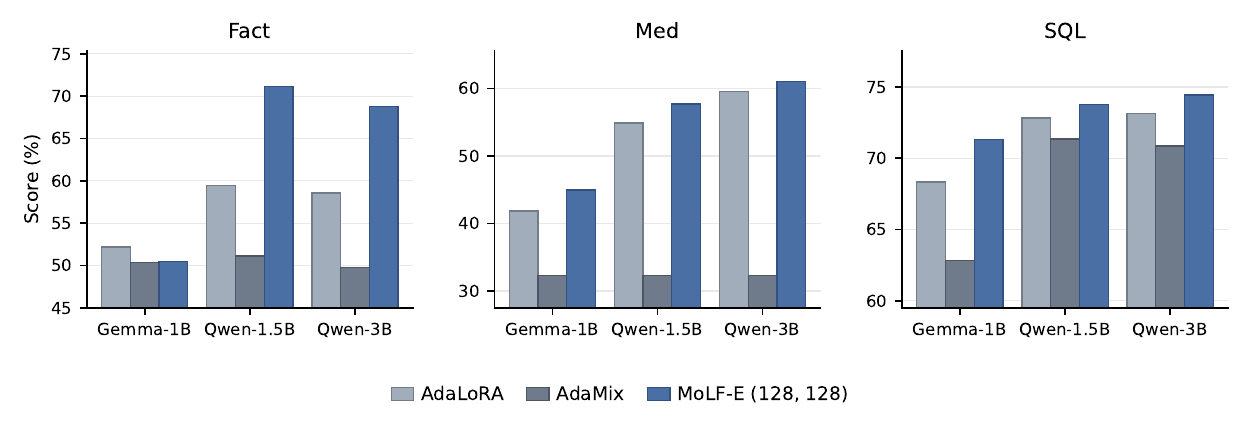}
\caption{MoLF-E with two rank-$128$ experts and Top-$1$ routing vs.\ adaptive PEFT baselines across three tasks and three models. MoLF-E (blue) scores higher than both adaptive baselines on $8$ of $9$ (task, model) settings.\looseness=-1}
\label{fig:method_compare}
\end{figure}

Figure~\ref{fig:method_compare} compares MoLF-E, using two independently initialized rank-$128$ LoRA experts and Top-1 routing, with AdaLoRA and AdaMix. MoLF-E scores higher than both in eight of nine settings; its largest margins over the stronger baseline are $11.70$ points on Fact, $3.13$ on Med, and $2.98$ on SQL. Appendix~\ref{subsec:ablation_rank} varies the first expert's rank.

\FloatBarrier

MoLF-E's advantage is largest on Fact. Independent low-rank experts may change optimization trajectories, but this comparison does not isolate that mechanism from capacity, tuning, or optimizer differences. As Section~\ref{sec:fft_vs_lft} shows, a uniquely heavy-tailed FFT endpoint does not explain the pattern. MoLF-E evaluates routing among low-rank candidates and has no dense update expert.

Appendix~\ref{app:efficiency} reports separate 7B/8B training-time and memory benchmarks. Standard MoLF retains dense optimizer states and incurs workload-dependent overhead; MoLF-E reduces memory but is not faster at the measured 7B operating point.

\subsection{Ablation of MoLF}
Table~\ref{tab:molf_ablation} ablates sparse updating ($K=1$ versus $K=2$) and the routing score (EPD versus PFN) on SQL and Med. Fact is examined through the routing trajectories in Figure~\ref{fig:router_evolve_main} and Appendix~\ref{subsec:router_behavior}.

\begin{table}[htbp]
\centering
\footnotesize 
\setlength{\tabcolsep}{3.5pt} 
\caption{Original ablations measured in accuracy (\%). Updating both experts ($K=2$) changes the update count as well as selection relative to Top-1 routing. EPD and PFN differ in their scoring functions; this comparison does not isolate the learning-rate factor alone.}
\label{tab:molf_ablation}
\begin{tabular}{lcccccc}
\toprule
\multirow{2}{*}{\textbf{Method}} & \multicolumn{2}{c}{\textbf{Qwen-3B}} & \multicolumn{2}{c}{\textbf{Qwen-1.5B}} & \multicolumn{2}{c}{\textbf{Gemma-1B}} \\
\cmidrule(lr){2-3} \cmidrule(lr){4-5} \cmidrule(lr){6-7}
& \textbf{SQL} & \textbf{Med} & \textbf{SQL} & \textbf{Med} & \textbf{SQL} & \textbf{Med} \\
\midrule
MoLF w/ no update routing  & 75.06 & 31.08 & 74.90 & 55.27 & 72.54 & 39.73 \\
MoLF w/ PFN-based routing & 75.03 & 59.14 & \textbf{74.92} & 55.77 & \textbf{73.09} & 42.31 \\
\midrule
\textbf{MoLF w/ EPD-based routing} & \textbf{75.41} & \textbf{60.60} & 74.44 & \textbf{56.28} & \textbf{73.09} & \textbf{45.40} \\
\bottomrule
\end{tabular}
\end{table}

\subsubsection{Sparse Update of Expert}
\label{subsubsec:ablation_select_all}

Updating both experts ($K=2$) lowers Med accuracy on Qwen2.5-3B ($60.60\%\to31.08\%$) and Gemma-3-1B ($45.40\%\to39.73\%$). EPD Top-1 improves all three Med configurations, while SQL differences are small and mixed. This comparison changes both selection and update count; its endpoint scores do not establish optimization oscillations, and the added physical updates' isolated runtime cost was not benchmarked.

Appendix~\ref{app:matched_routing} compares EPD with a coin-flip Top-1 selector at the same update count over five paired seeds, and Appendix~\ref{app:projection_routing} reports projection-level selection frequencies. These additional results are scoped to the measured settings, with the control's seed-44 collapse and its influence on the \mbox{mean reported explicitly.}

\subsubsection{Expert Selection Heuristics}
\label{subsubsec:ablation_expert_selection}


We compare EPD with an author-constructed Preconditioned Frobenius Norm (PFN) diagnostic: the RMS preconditioned Adam direction, which approximately cancels uniform gradient scaling when $\epsilon$ is negligible (Appendix~\ref{app:pfn_details}).

EPD matches or improves over PFN in five of six cells, with Med gains of approximately $3.1$ points on Gemma-3-1B and $1.46$ on Qwen2.5-3B; it trails PFN by $0.48$ points on Qwen2.5-1.5B SQL. These results suggest a setting-dependent benefit from EPD's additional scoring information. They neither isolate its learning-rate factor nor test rank-invariant routing among \mbox{unequal-rank MoLF-E experts.}

\begin{figure}[htbp]
\centering
\includegraphics[width=\linewidth]{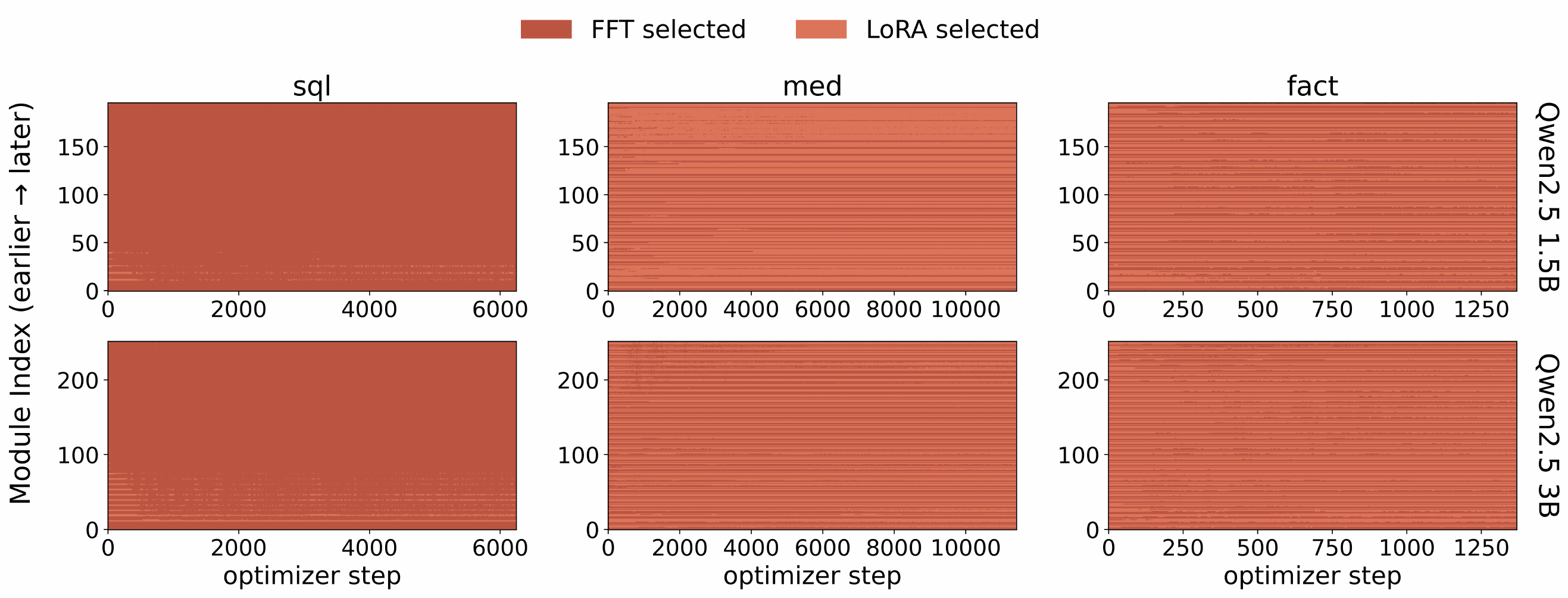}
\caption{MoLF routing dynamics over training. Each heatmap row tracks one module's structural preference (FFT vs.\ LoRA) across optimizer steps. The distinct horizontal striping shows routing stabilizes rapidly: modules commit early to either dense or low-rank experts with minimal oscillation.}
\label{fig:router_evolve_main}
\end{figure}

Figure~\ref{fig:router_evolve_main} shows persistent module-level preferences with a smaller set of switching modules. The mixture varies by task and backbone, without a universal LoRA-first/FFT-later pattern. These trajectories describe routing behavior rather than establish a performance mechanism; Appendix~\ref{subsec:router_behavior} gives the full trajectories.
 
\section{Conclusion}
\label{sec:conclusion}
Fine-tuning pretrained LLMs involves a task-dependent balance between FFT's dense capacity and LoRA's potential implicit regularization. In our tested settings, FFT leads on CounterFact, tuned higher-rank LoRA on MedMCQA, and both perform similarly on Text-to-SQL. Greater capacity is therefore not uniformly beneficial: low-rank constraints may regularize adaptation in some settings while limiting task-relevant updates in others. Endpoint compressibility alone does not explain these differences; task sensitivity and optimization trajectories remain possible explanations. MoLF addresses this choice through module-local routing between dense and low-rank updates. Its original nine-setting point estimates are above or within $1.5$ percentage points of the better tuned baseline; LoRA-only MoLF-E exceeds both adaptive LoRA baselines in eight of nine settings.

The five-seed CounterFact study finds a higher MoLF mean and lower sample variability, although the paired MoLF--FFT interval includes zero and learning rates differ. The 8B reasoning experiment shows an advantage over matched LoRA, with an inconclusive difference from FFT under evaluation-problem uncertainty. Standard MoLF retains dense training-state costs and workload-dependent time overhead; MoLF-E targets lower-memory settings.

Future work includes larger candidate pools, which may enable finer-grained adaptation but increase gradient, optimizer-state, and computation costs. Higher-order information or explicit rank calibration could improve the first-order EPD heuristic. MoLF-E trades full-rank expressiveness for lower memory, motivating combinations with more expressive efficient-LoRA methods. Systems optimization, coordinate-sparse comparisons, and validating early routing as a diagnostic for static-method selection also merit study. Appendix~\ref{sec:limitations} details the empirical, mechanistic, and practical limitations.

\begin{ack}
We have no additional funding or competing interests to declare beyond those associated with our stated affiliations.

\end{ack}

\pagebreak
\bibliography{references}






\newpage
\appendix

\section{Limitations}
\label{sec:limitations}

\paragraph{Statistical and evaluation scope.}
Each method's result in the original nine tested settings comes from one training run.
The additional five-seed CounterFact study provides evidence for one setting,
not seed robustness across the full table. In the tested settings,
FFT uses a different learning rate from MoLF, and the LoRA control has its own
limited tuning grid. MoLF's higher mean and lower observed seed variability
are descriptive; the paired MoLF--FFT interval includes zero
(Appendix~\ref{app:seed_study}). The 8B reasoning comparison uses one training
seed per configuration and three generation seeds. Its bootstrap intervals
quantify evaluation-problem variation conditional on those trained models,
not variation across training runs. A MoLF--FFT interval containing zero does
not establish equivalence. Frequent generation truncation, particularly for
LoRA, further limits the interpretation of MATH-500 differences
(Appendix~\ref{app:math_protocol}). Broader reasoning tasks, larger backbones,
and benchmarks without a corresponding supervised training set remain to be
studied. The random-routing control is also limited to one model--task setting;
one unusually weak random-routing seed is insufficient to establish a general
stability advantage (Appendix~\ref{app:matched_routing}).

\paragraph{Training cost and the scope of MoLF-E.}
Standard MoLF retains trainable dense weights, dense gradients, and dense Adam
states in addition to its LoRA pathway. Top-1 routing skips a pathway's
physical parameter update, but does not skip its forward/backward computation
or moment tracking. Standard MoLF therefore does not provide LoRA-like
training-memory savings. The measured time and memory costs depend on workload,
precision, optimizer implementation, and distributed execution; the 7B
short-context and 8B context-sweep measurements should not be combined into a
universal overhead estimate (Appendix~\ref{app:efficiency}). MoLF-E reduces
persistent state by freezing the backbone, but consequently cannot select a
dense FFT update or retain standard MoLF's full-rank expressiveness. Systems
optimization of the current implementation remains future work.

\paragraph{Mechanistic interpretation.}
The spectral analysis uses one selected Qwen2.5-1.5B FFT endpoint per task.
Its layer-bootstrap intervals describe variation within those endpoints,
rather than training-seed uncertainty. Fact is the most compressible endpoint
by spectral tail energy, yet has the largest absolute rank-128 truncation
loss; normalizing that loss by total FFT gain reverses the ordering. These
results motivate loss sensitivity and constrained optimization trajectories
as explanations, but do not establish either causally. Endpoint truncation
is not a trained LoRA solution. Likewise, capability preservation on Med and
low-dimensional adaptation on SQL remain hypotheses rather than measured
mechanisms (Appendix~\ref{app:spectral}).

\paragraph{Score calibration and heterogeneous ranks.}
EPD is a heuristic for descent per trainable parameter, not a
parameterization-invariant estimate of an expert's utility. Its value depends
on the configured learning rate, parameter count, and LoRA scaling. Under the
simplifying gradient-scaling assumptions, RS-LoRA gives an approximate score
dependence of $\eta_i\alpha_i/\sqrt{r_i}$; thus, the MoLF-E rank sweep changes
both representational capacity and routing preference. The fixed-rank PFN
control does not test rank-invariant routing in MoLF-E. Rank-invariant scores,
rank-aware learning rates, and routing priors require direct comparisons.

\paragraph{Candidate families and scaling.}
The experiments focus on the minimal dense/LoRA pair and on a pair of LoRA
pathways for MoLF-E. Every added candidate participates in computation and
retains optimizer moments, so increasing the candidate pool increases cost.
The unequal-rank MoLF-E sweep and the two-candidate $K=1/2$ ablation do not
establish the benefit of larger pools, rank-calibrated routing, choices of
$K$ beyond that pair, or load balancing. Coordinate-sparse fine-tuning constrains update
support rather than matrix rank and is a complementary candidate, but we do
not have a completed matched FFT/sparse-fine-tuning routing comparison.
Consequently, these results do not show that LoRA is uniquely the best
constrained counterpart to FFT.

\paragraph{Early routing as a diagnostic.}
Routing often stabilizes early, but we have not established that early EPD
scores predict which static method will achieve the best final performance.
EPD requires observed gradients and nonzero moment estimates, so it cannot
make this decision before training. A short burn-in followed by static
FFT/LoRA selection would require a validated aggregation rule and horizon,
and would give up MoLF's subsequent module-local adaptation.

\section{Experimental Details}
\label{sec:appendix_experimental_details}

\subsection{Derivation of the Expected Preconditioned Descent (EPD) Score}
\label{sec:epd_derivation}

The Expected Preconditioned Descent (EPD) score ($\mathcal{S}_t^{(i)}$) introduced in Equation~\ref{eq:epd_score} is motivated by a first-order heuristic for loss reduction per parameter. The approximation below omits weight decay and higher-order terms and is not a guarantee of descent or an unbiased estimator of a specified probabilistic expectation.\looseness=-1

By first-order Taylor expansion, the expected change in the loss function $\mathcal{L}$ after an optimization step $\Delta \theta$ is approximated by the inner product of the gradient and the step direction:
\begin{equation}
\Delta \mathcal{L} \approx (\nabla_\theta \mathcal{L})^\top \Delta \theta .
\label{eq:taylor}
\end{equation}

For the AdamW optimizer (omitting decoupled weight decay for the heuristic), the parameter update direction is preconditioned by its moving averages:
\begin{equation}
\Delta \theta_t \approx -\eta_t \frac{m_t}{\sqrt{v_t} + \epsilon}.
\label{eq:preconditioned}
\end{equation}

Under the additional approximation that the current gradient is represented by its first moment ($g_t \approx m_t$), substituting the preconditioned direction from Equation~\ref{eq:preconditioned} into the Taylor expansion in Equation~\ref{eq:taylor} motivates the loss-reduction proxy. This approximation is not implied by smoothness alone and can fail when gradients change rapidly:
\begin{equation}
-\Delta \mathcal{L} \approx g_t^\top \left( \eta_t \frac{m_t}{\sqrt{v_t} + \epsilon} \right) \approx \eta_t \sum_{\theta} \frac{m_t^2}{\sqrt{v_t} + \epsilon} 
\label{eq:expected_loss}
\end{equation}

We normalize the proxy by the parameter count $N_{\text{params}}^{(i)}$ of expert $i$ to compare predicted descent per parameter. This is a design choice that makes candidate size part of the selection criterion, rather than a guarantee of rank or parameterization invariance. It yields the EPD score used for dynamic routing:\looseness=-1
\begin{equation}
\mathcal{S}_t^{(i)} = \frac{\eta_t^{(i)}}{N_{\text{params}}^{(i)}} \sum_{\theta \in \Theta_i} \frac{\left(m_t^{(i)}\right)^2}{\sqrt{v_t^{(i)}} + \epsilon} 
\label{eq:epd_score_appendix}
\end{equation}

\subsection{Baseline Hyperparameter Sweep}
\label{subsec:hyperparam_sweep}
To establish rigorous baselines, we conduct a comprehensive hyperparameter sweep for each model architecture (Qwen2.5-1.5B, Qwen2.5-3B, and Gemma-3-1B) across all three benchmark datasets (SQL, Fact, and Med). We optimize the configurations for both Low-Rank Adaptation (LoRA) and Full Fine-Tuning (FFT) to ensure the best possible performance for our baselines. 

The search spaces for both tuning methods are detailed in Table~\ref{tab:hyperparam_sweep}. For all configurations, we maintain a fixed warmup ratio of $0.1$.

\begin{table}[htbp]
\centering
\caption{Hyperparameter search space for LoRA and Full Fine-Tuning (FFT) baselines.}
\label{tab:hyperparam_sweep}
\begin{tabularx}{\textwidth}{l >{\centering\arraybackslash}X >{\centering\arraybackslash}X c >{\centering\arraybackslash}X}
\toprule
\textbf{Method} & \textbf{Learning Rate} & \textbf{Scheduler} & \textbf{Warmup Ratio} & \textbf{Rank ($r$)} \\
\midrule
LoRA & \num{5e-4}, \num{5e-5} & Cosine, Linear & $0.1$ & $8, 16, 32, 64, 128$ \\
FFT & \num{1e-4}, \num{5e-5}, \num{2e-5} & Cosine, Linear & $0.1$ & -- \\
\bottomrule
\end{tabularx}
\end{table}

\paragraph{MoLF and MoLF-E hyperparameters.} In the original nine model--task comparisons, the reported MoLF configuration pairs the trainable base weights with one rank-$128$ LoRA expert. We fix the LoRA-expert learning rate at \num{5e-5} and select the best EPD result over base-expert learning rates $\{\num{1e-4},\num{5e-5},\num{2e-5}\}$ and linear and cosine schedulers. The reported MoLF-E configuration uses two independently initialized rank-$128$ LoRA experts, fixes both learning rates at \num{5e-5}, and selects the better result from the linear and cosine schedulers.
The per-expert decoupled weight decay $\lambda_i$ in Equation~\ref{eq:topk_update} is $\lambda_{\text{FFT}} = 0.1$ for the FFT expert and $\lambda_{\text{LoRA}} = 0.01$ for every LoRA expert. The stability constant $\epsilon$ is set to the HuggingFace AdamW default ($\epsilon = 10^{-8}$). For both MoLF and MoLF-E we use $K = 1$ for the Top-$K$ routing in Phase~3, so exactly one expert per module receives a physical weight update at each step. These original runs use a warmup ratio of $0.1$, and AdamW uses $(\beta_1, \beta_2) = (0.9, 0.999)$. The additional experiments have separately specified protocols in Appendix~\ref{app:additional_experiments} and Appendix~\ref{app:routing_controls}.

\paragraph{AdaLoRA and AdaMix configuration.} For AdaLoRA~\citep{zhang2023adaptive} we follow the rank-pruning schedule of the original paper and reuse the best LoRA learning rate found in the sweep above ($5 \times 10^{-4}$) across all (model, task) configurations, with a single exception: on Gemma-3-1B Fact the learning rate is increased to $5 \times 10^{-3}$ to obtain reasonable performance. We additionally tried a learning rate of $5 \times 10^{-5}$ at a higher initial rank ($r = 128$), but this configuration consistently degraded accuracy and is not reported. For AdaMix~\citep{wang2022adamix} we evaluated both the standard hyperparameters recommended by the original paper and a learning rate of $5 \times 10^{-5}$; the latter outperformed the former in our setup and is the configuration reported in Section~\ref{subsec:molf_efficient}.

\subsection{Preconditioned Frobenius Norm (PFN) Formulation}
\label{app:pfn_details}

In Section~\ref{subsubsec:ablation_expert_selection}, we use the Preconditioned Frobenius Norm (PFN) as an author-constructed diagnostic directional-consistency control. It is the root-mean-square of the preconditioned Adam direction:\looseness=-1

\begin{equation}
\mathcal{S}_{\text{PFN}}^{(i)} = \frac{1}{\sqrt{N_{\text{params}}^{(i)}}} \sqrt{\sum_{\theta \in \Theta_i} \left( \frac{m_t^{(i)}}{\sqrt{v_t^{(i)}} + \epsilon} \right)^2}
\end{equation}

Under a uniform rescaling of the gradient history, $m_t$ and $\sqrt{v_t}$ rescale together, so their ratio approximately cancels the scale when $\epsilon$ is negligible. PFN therefore tests whether a normalized consistency signal is sufficient for selection, whereas EPD retains gradient-magnitude and learning-rate information. Both scores normalize for parameter count, in different ways. PFN's approximate scale invariance is not invariance to arbitrary changes in LoRA parameterization. Moreover, the existing PFN experiment compares FFT with one fixed-rank LoRA candidate; it does not test rank-invariant routing among unequal-rank MoLF-E candidates. Rank-invariant EPD variants, rank-aware learning rates, and routing priors remain untested alternatives.

\subsection{Evaluation Metrics}
\label{subsec:metrics}

We report two evaluation metrics across the three benchmarks. On Med (MedMCQA) and SQL (Gretel synthetic Text-to-SQL) we report standard accuracy: 4-way multiple-choice accuracy on the held-out MedMCQA validation split, and exact-match accuracy on the held-out Text-to-SQL queries. On Fact (CounterFact) we report the Efficacy Score (ES) introduced by~\citet{meng2022locating}, which measures the fraction of edits for which the post-edit model assigns higher probability to the new counterfactual target than to the original true object:
\begin{equation}
\text{ES} = \frac{1}{|\mathcal{E}|} \sum_{(\pi, o^*, o^c) \in \mathcal{E}} \mathbb{1}\!\left[\, \mathbb{P}_{G'}(o^* \mid \pi) > \mathbb{P}_{G'}(o^c \mid \pi) \,\right],
\label{eq:es}
\end{equation}
where $G'$ is the fine-tuned (edited) model, $\pi$ is the counterfactual prompt, $o^*$ is the counterfactual target object, $o^c$ is the original true object, $\mathcal{E}$ is the set of edits, and $\mathbb{1}[\cdot]$ is the indicator function. We report ES as a percentage so that all three benchmarks share a common $[0, 100]$ \mbox{scale in our tables and figures.}

\subsection{Compute Resources}
\label{subsec:compute}

The original nine-setting experiments were executed on a single NVIDIA H100 Tensor Core GPU or a single NVIDIA RTX PRO 6000 Blackwell GPU, with no model or data parallelism. A run in that campaign took approximately 2 to 6 hours of wall-clock time, depending on the (model, task) configuration: larger backbones (Qwen2.5-3B) and longer-context tasks (SQL) were at the upper end, whereas smaller backbones (Gemma-3-1B) and shorter-context tasks (Med, Fact) were at the lower end. The original campaign, including the full learning-rate, scheduler, and LoRA-rank sweeps, ablations, and unreported configurations, was estimated at approximately 2500 H100 GPU-hours. This rough estimate is not a revised aggregate including the additional experiments below. Appendix~\ref{app:efficiency} reports the separate single-H100 7B and eight-H100 8B measurement protocols; these workload-specific measurements should not be extrapolated to a total project cost.

\section{FFT Endpoint Spectra and Task-Score Truncation}
\label{app:spectral}

\subsection{Checkpoint provenance and analysis scope}
\label{app:spectral_provenance}
We study displacements from the pretrained, non-instruction-tuned Qwen2.5-1.5B model to the archived FFT endpoints selected by the original sweep. Table~\ref{tab:spectral_endpoints} identifies the tested settings and their archived scores. The endpoint recovery and validation record identifies the hashed base snapshot by revision \texttt{8faed761\ldots}. All three archived endpoints were recovered and validated; the results reported here use those endpoints rather than the separately produced fallback reruns. There is one selected endpoint per task, and their training seeds were not recovered.

\begin{table}[ht]
\centering
\caption{Archived FFT endpoints used for spectral analysis. All use a linear schedule, a warmup ratio of 0.1, and AdamW weight decay of 0.1. Scores are Efficacy Score (\%) for Fact and accuracy (\%) for Med and SQL.\looseness=-1}
\label{tab:spectral_endpoints}
\begin{tabular}{@{}lcc@{}}
\toprule
Task & Learning rate & Archived selection score \\
\midrule
Fact & $10^{-4}$ & 74.948675 \\
Med & $5\times10^{-5}$ & 55.390868 \\
SQL & $10^{-4}$ & 74.812693 \\
\bottomrule
\end{tabular}
\end{table}

For each task we analyze the same seven projection families, \texttt{q\_proj}, \texttt{k\_proj}, \texttt{v\_proj}, \texttt{o\_proj}, \texttt{gate\_proj}, \texttt{up\_proj}, and \texttt{down\_proj}, in all 28 transformer blocks. This gives 196 matrices per task and 588 matrices in total. Embeddings, the language-model head, normalization parameters, and biases are excluded from this spectral matrix set. Matrix keys and shapes are matched to the base checkpoint. The q/o matrices have shape $1536\times1536$, k/v have shape $256\times1536$, gate/up have shape $8960\times1536$, and down has shape $1536\times8960$. The summaries give each analyzed matrix equal weight; they are not parameter-count-weighted summaries.

\subsection{Definitions, numerical checks, and layer bootstrap}
\label{app:spectral_protocol}
For task $t$ and matrix $j$, define the endpoint displacement
\begin{equation}
    D_{t,j}=W^{\mathrm{FFT}}_{t,j}-W^0_j.
\end{equation}
Base and endpoint tensors are cast separately to FP32 before subtraction, avoiding subtraction in BF16. We compute all singular values with exact \texttt{torch.linalg.svdvals} and accumulate squared energies in FP64. For sorted singular values $\sigma_1\ge\cdots\ge\sigma_q$, the relative rank-$r$ tail energy is
\begin{equation}
    T_r(D)=\frac{\sum_{i>r}\sigma_i^2}{\sum_i\sigma_i^2}
    =\frac{\min_{\operatorname{rank}(X)\le r}\|D-X\|_F^2}{\|D\|_F^2}.
    \label{eq:spectral_tail}
\end{equation}
The equality follows from Eckart--Young--Mirsky~\citep{eckart1936approximation,mirsky1960symmetric}. It is a parameter-space approximation bound, not a task-loss bound. Tail energy is computed by directly summing the discarded squared singular values. We also record relative displacement $\|D\|_F/\|W^0\|_F$ and stable rank $\|D\|_F^2/\|D\|_2^2$.

The computation checks finite tensors and singular values, a relative discrepancy of at most $10^{-3}$ between the SVD energy and the directly computed squared Frobenius norm, $T_r\in[0,1]$ within $10^{-5}$, and monotonicity over $r\in\{8,16,32,64,128\}$. A zero-energy update would make $T_r$ undefined and invalidate that paired primary observation; it would not be silently assigned zero. All 588 archived rows passed, with no zero-energy updates or numerical failures, and all 196 Fact/SQL pairs were valid.\looseness=-1

The predeclared primary question was whether Fact has more energy beyond rank 128 than SQL. Its statistic is the median of the \emph{paired differences},
\begin{equation}
    \delta_j=T_{128}(D_{\mathrm{Fact},j})-T_{128}(D_{\mathrm{SQL},j}),
    \qquad
    \Delta_{128}=\operatorname{median}_{j=1}^{196}\delta_j.
    \label{eq:spectral_paired}
\end{equation}
This differs from subtracting the two task medians. We use 10,000 percentile cluster-bootstrap draws with random seed 42. Each draw samples 28 layer IDs with replacement, retaining all seven projection families and their Fact/SQL pairing within each sampled layer, and then recomputes $\Delta_{128}$. Its 2.5th and 97.5th percentiles form the reported interval. This resampling preserves dependence among projection families in a layer. It measures layer heterogeneity for one selected endpoint pair; it does not measure training-seed uncertainty or generalization to other tasks or models.

\subsection{Archived endpoint diagnostics}
\label{app:spectral_diagnostics}
Table~\ref{tab:spectral_diagnostics} reports the available archived task summaries. Fact has the smallest median tail energy, effective rank, and stable rank. The primary paired result is
\begin{equation}
    \Delta_{128}=-0.109298,
    \qquad 95\%\ \text{layer-cluster interval}=[-0.126566,-0.098964].
\end{equation}
It does not support the proposed ordering in which Fact has a heavier rank-128 endpoint tail than SQL. This result concerns the selected FFT displacements; it does not establish that Fact has lower task-intrinsic dimension.

\begin{table}[ht]
\centering
\caption{Descriptive archived FFT displacement diagnostics for Qwen2.5-1.5B, summarized over 196 matrices per task. The base-subspace quantity is the reported fraction of update energy in the dominant rank-128 pretrained subspace.}
\label{tab:spectral_diagnostics}
\begin{tabular}{@{}lrrr@{}}
\toprule
Diagnostic & Fact & Med & SQL \\
\midrule
Median $T_{128}$ & 0.450 & 0.639 & 0.598 \\
Effective rank of $D$ & 446 & 778 & 730 \\
Stable rank of $D$ & 20.9 & 44.6 & 70.4 \\
Relative displacement $\|D\|_F/\|W^0\|_F$ & 0.0338 & 0.0281 & 0.0648 \\
Energy in base top-128 subspace & 0.0061 & 0.0057 & 0.0068 \\
\bottomrule
\end{tabular}
\end{table}

The similarly small base-subspace energy fractions across all three tasks do not distinguish Med from Fact or SQL, so displacement outside dominant pretrained directions alone cannot explain selective capability loss on Med. This is also not a direct forgetting measurement: retained capabilities require evaluation on suitable tasks.

An exploratory analysis of resulting weight matrices, rather than displacements, covered the three original model sizes, three tasks, and both FFT and rank-128 LoRA. The reported effective-rank ratios relative to the base span 1.00003--1.00187, changes of at most about 0.19\% against effective ranks of 392--871. These small changes make this diagnostic insensitive to the proposed distinction that LoRA adds rank while FFT perturbs existing directions. The LoRA-minus-FFT rank differences were at most approximately one rank unit and were confounded by the selected learning rates; only Med used matched rates. We therefore do not interpret this exploratory pattern as evidence for a method-specific mechanism.

\subsection{Task-score effects of endpoint truncation}
\label{app:spectral_truncation}
For the analyzed matrices, we form rank-$r$ approximations to the FFT displacements and evaluate the corresponding truncated checkpoints:
\begin{equation}
    W^{(r)}_j=W^0_j+\operatorname{SVD}_r(D_j).
\end{equation}
This intervention tests the task-score effect of discarding endpoint directions. It is not training a rank-$r$ LoRA adapter: the SVD is taken after FFT has completed. Table~\ref{tab:spectral_truncation} gives the reported base, truncated, and full-endpoint scores. The full-endpoint scores are those of this truncation evaluation, which differ slightly from the archived selection scores in Table~\ref{tab:spectral_endpoints}; the gaps and normalizations use the scores within the truncation evaluation consistently.

\begin{table}[ht]
\centering
\caption{Task-score effects of truncating the selected FFT endpoint displacement. Scores are Efficacy Score (\%) on Fact and accuracy (\%) on Med/SQL. Gap@128 is full minus rank-128 score; FFT gain is full minus base score. No training-seed uncertainty is estimated for this endpoint intervention.}
\label{tab:spectral_truncation}
\resizebox{\textwidth}{!}{%
\begin{tabular}{@{}lrrrrrrrr@{}}
\toprule
Task & Base & Rank 8 & Rank 32 & Rank 128 & Full FFT & Gap@128 & FFT gain & Gap/gain \\
\midrule
Fact & 16.87 & 58.36 & 64.77 & 70.04 & 74.96 & 4.92 & 58.09 & 8.5\% \\
Med & 45.40 & 50.25 & 52.62 & 53.98 & 55.39 & 1.41 & 9.99 & 14.1\% \\
SQL & 63.22 & 47.18 & 66.13 & 73.07 & 74.77 & 1.70 & 11.55 & 14.7\% \\
\bottomrule
\end{tabular}%
}
\end{table}

Rank-128 truncation costs 4.92 absolute task-score points on Fact, compared with 1.41 on Med and 1.70 on SQL, despite Fact's smaller Frobenius tail. However, Fact also has a much larger total FFT gain over the base model. Dividing each gap by its total FFT gain reverses the ordering: 8.5\% on Fact versus 14.1\% on Med and 14.7\% on SQL. Both views are necessary to interpret the task comparison. SQL's rank-8 truncated endpoint scores below its base model, showing that sufficiently aggressive endpoint truncation can damage performance rather than merely remove the fine-tuning gain.

\subsection{Interpretation and limits}
\label{app:spectral_limits}
The evidence rules out using the observed endpoint-tail ordering to explain Fact's FFT advantage. It instead motivates two hypotheses. First, the task score may be sensitive to relatively low-energy discarded directions, so parameter-space approximation quality alone is insufficient. Second, optimizing low-rank factors constrains the trajectory at every step, whereas post-hoc SVD constrains only the final displacement. The factor subspaces evolve during LoRA training; the hypothesis does not require a fixed subspace or imply that a compressible FFT endpoint is unreachable by LoRA. These experiments do not separate these explanations causally, and a matched comparison of trained LoRA with truncated FFT at each rank would be needed to investigate their relative contributions.

The endpoints were selected by task score, rather than replicated under a common random-seed protocol. Fact and SQL share the listed learning rate and schedule but differ in data, objectives, and optimizer-step counts; Med uses a different learning rate. Update scale also differs across tasks, as shown in Table~\ref{tab:spectral_diagnostics}. AdamW weight decay can contribute a full-rank displacement aligned with the pretrained weights, and BF16 checkpoint storage can erase small changes even when subtraction is later performed in FP32. These factors limit causal interpretation of the spectra. The layer bootstrap does not remove these limitations. We therefore describe SQL as empirically rank-insensitive in the sweep, treat the Med capability-preservation explanation as a hypothesis, and interpret the Fact result through measured score sensitivity rather than an asserted heavy-tailed optimal update.

\subsection{Rank constraints and regularization hypotheses}
\label{app:rank_mechanisms}
The task interpretations in Section~\ref{sec:fft_vs_lft} distinguish parameter-space approximation from task performance. Let $W_{\mathrm{base}}\in\mathbb{R}^{d\times k}$ denote the pretrained weights and let $\Delta W^*=\sum_{i=1}^{\min(d,k)}\sigma_i u_i v_i^\top$ be a candidate target update. By the Eckart--Young--Mirsky theorem~\citep{eckart1936approximation,mirsky1960symmetric}, any rank-$r$ \mbox{approximation $BA$ satisfies}
\begin{equation}
    \|\Delta W^*-BA\|_F^2\geq\sum_{i>r}\sigma_i^2.
\end{equation}
A heavy spectral tail would imply a large parameter-space representation error for this target; a concentrated spectrum would permit a small residual. This conditional argument does not measure the unknown task-optimal update, bound task loss, predict monotone improvement with trained LoRA rank, or establish task-intrinsic dimension. In particular, the measured Fact endpoint is more Frobenius-compressible than the Med and SQL endpoints, contrary to a heavier-tail explanation for its FFT advantage. Discarded directions may nevertheless be task-sensitive, and training low-rank factors constrains the trajectory rather than only the final displacement.

For the Med regularization hypothesis, suppose the empirical gradient decomposes as $G=G_{\mathrm{task}}+G_{\mathrm{noise}}$, with a high-rank noise component. Under vanilla SGD on unscaled LoRA factors, simultaneous updates are $B_{\mathrm{new}}=B-\eta GA^\top$ and $A_{\mathrm{new}}=A-\eta B^\top G$, giving
\begin{equation}
    B_{\mathrm{new}}A_{\mathrm{new}}-BA
    =-\eta\bigl(BB^\top G+GA^\top A\bigr)+\mathcal{O}(\eta^2).
\end{equation}
Each first-order term has rank at most $r$ and acts through the current column space of $B$ or row space of $A$; these spaces evolve during training. Even with orthonormal factor columns/rows, the sum of left and right projections is not itself an orthogonal projection: components in their intersection are counted twice. General factors also rescale the update through their singular values. This geometry can attenuate gradient components but does not prove that retained components are signal. AdamW's adaptive preconditioning further modifies the factor updates. Thus the argument supplies a regularization intuition, while reduced forgetting remains unestablished without direct pretrained-capability evaluations.

\section{Additional Performance and Efficiency Experiments}
\label{app:additional_experiments}

\subsection{Five-seed CounterFact evaluation}
\label{app:seed_study}

To assess the small CounterFact margin in the original comparison, Table~\ref{tab:counterfact_seed_summary} summarizes five paired training seeds on Qwen2.5-1.5B. We report mean and sample SD, with differences computed seed by seed. In these tested settings, FFT uses $10^{-4}$, while both MoLF pathways and the selected LoRA-128 control use $5\times10^{-5}$. The following protocols, per-seed results, and selection \mbox{analysis qualify this comparison.}

\begin{table}[htbp]
\centering
\caption{Qwen2.5-1.5B CounterFact over five paired training seeds. ES is reported as mean $\pm$ sample SD; the paired difference is computed seed by seed.}
\label{tab:counterfact_seed_summary}
\begin{tabular}{lcc}
\toprule
Method & ES (\%) & MoLF $-$ method (points) \\
\midrule
FFT & $71.043 \pm 2.579$ & $+3.523 \pm 3.031$ \\
LoRA-128 & $70.346 \pm 0.491$ & $+4.220 \pm 0.945$ \\
MoLF & $74.566 \pm 0.872$ & --- \\
\bottomrule
\end{tabular}
\end{table}

\paragraph{Training and evaluation protocol.}
We evaluate Qwen2.5-1.5B on the full CounterFact dataset of 21,919 examples
using five paired training seeds, 42--46. All methods share the same fixed
sample order within each seed. Training uses two epochs, microbatch size 8,
four-step gradient accumulation (effective batch size 32), and 1,370
optimizer steps. We use a linear schedule with warmup ratio 0.1, AdamW
$(\beta_1,\beta_2)=(0.9,0.999)$ and $\epsilon=10^{-8}$, maximum gradient norm
1.0, BF16 training, TF32, and scaled dot-product attention. The fixed ES
evaluator uses BF16, batch size 48, greedy generation with at most 32 new
tokens, and evaluation seed 42.

FFT uses learning rate $10^{-4}$ and weight decay 0.1. MoLF wraps all seven
projection families in 28 layers (196 modules), with one dense and one
rank-128 RS-LoRA pathway in each module. Both MoLF pathways use learning rate
$5\times10^{-5}$, with dense/LoRA weight decay $0.1/0.01$, LoRA scaling
constant $\alpha=32$, dropout 0.05, and Top-1 EPD routing. These are additional
runs with explicitly specified adapter settings, rather than bitwise
reproductions of the original results table.

\begin{table}[htbp]
\centering
\small
\caption{Five paired training seeds on Qwen2.5-1.5B CounterFact. ES is in
percent and differences are in percentage points; SD denotes sample standard deviation.
FFT and MoLF use different selected learning rates.\looseness=-1}
\label{tab:counterfact_seed_details}
\begin{tabular}{lrrr}
\toprule
Seed & MoLF ES (\%) & FFT ES (\%) & MoLF $-$ FFT \\
\midrule
42 & 75.291 & 70.993 & $+4.298$ \\
43 & 75.496 & 69.319 & $+6.177$ \\
44 & 73.598 & 75.200 & $-1.601$ \\
45 & 73.735 & 68.525 & $+5.210$ \\
46 & 74.711 & 71.180 & $+3.531$ \\
\midrule
Mean $\pm$ SD & $74.566\pm0.872$ & $71.043\pm2.579$ & $+3.523\pm3.031$ \\
\bottomrule
\end{tabular}
\end{table}

MoLF has higher ES in four of five paired seeds. Its observed mean advantage
is $3.523$ points, with nominal 95\% paired-$t$ interval
$[-0.241,+7.287]$. This interval includes zero, so the original small
advantage does not establish a statistically supported improvement.
Its smaller sample SD describes these five runs rather than establishing a
general variance-reduction property. Because FFT uses
$10^{-4}$ whereas MoLF uses $5\times10^{-5}$, this is a comparison of
tested method/configuration pairs, not an experiment isolating the routing
method at a matched learning rate. The original margin of about
$0.205$ points should be interpreted in light of this training-seed variation.

\paragraph{Additional LoRA control and selection procedure.}
The additional LoRA search evaluates ranks
$\{8,16,32,64,128,256\}$ and learning rates
$\{5\times10^{-4},5\times10^{-5}\}$. A two-seed pilot (42--43) selects the
higher-mean learning rate separately for each rank; the resulting configuration
is then evaluated on seeds 44--46. This entails 24 pilot runs and 18
additional runs. The adapters use RS-LoRA, dropout 0.05, weight decay 0.01,
and $\alpha=\operatorname{round}(2\sqrt{r})$. These adapter scales differ
from the fixed \mbox{$\alpha=32$ used by MoLF.}

Among the ranks at most 128, the highest five-seed mean is achieved by
LoRA-128 at learning rate $5\times10^{-5}$:
$70.346\pm0.491$ ES. Its mean over the three seeds not used for learning-rate
selection is $70.202\pm0.276$. MoLF exceeds this control in all five paired
seeds, with mean difference $4.220\pm0.945$ points. The corresponding nominal
paired-$t$ interval is $[3.047,5.394]$; with only five seeds and configuration
selection, we treat this as a descriptive targeted comparison. LoRA-128 minus
FFT is $-0.697\pm2.451$ points, with LoRA higher in three of five seeds.

The two-point LoRA learning-rate grid is coarse, and the five-seed means
include the two seeds used for learning-rate selection. Neither this search
nor the earlier MoLF search establishes an equally tuned optimum across
methods. Rank 256 is outside the low-rank comparison because the
$256\times1536$ key/value matrices then admit full-rank adapter updates;
for transparency, its five-seed ES is $74.586\pm0.874$, close to MoLF's
$74.566\pm0.872$. This observation is compatible with a capacity contribution
to the rank-at-most-128 gap, rather than showing that routing alone is
necessary for the CounterFact improvement.

\FloatBarrier
\subsection{8B mathematical-reasoning evaluation}
\label{app:math_protocol}

\paragraph{Data, model, and training.}
We use Meta-Llama-3-8B with its native 8,192-token context and no RoPE
extension. From ODA-Mixture-100k, we retain the 34,702 examples whose tokenized
length is at most 8,192 and drop the remaining 66,604 rather than truncating
their contents. Best-fit packing yields 25,280 bins, with 10.67\% unused
token positions and a reported supervised-token fraction of 87.2\%.
Each configuration trains for one epoch, or 790 optimizer steps, with
training seed 42. The primary comparison uses the same peak learning rate
$5\times10^{-5}$ for FFT, rank-128 LoRA, and both MoLF pathways.\looseness=-1

\paragraph{Evaluation and uncertainty.}
We evaluate all 500 MATH-500 problems using the zero-shot
\texttt{Human:}/\texttt{Assistant:} prompt format used in training, with the
\texttt{jet-gen} generation engine and xVerify judging. The maximum is
7,168 generated tokens within the native 8,192-token window. Accuracy
averages three generation seeds for each trained model. Let $Y_{m,i,z}$ be
the correctness indicator for method $m$,
problem $i$, and generation seed $z$. We compute
\begin{equation}
\bar{Y}_{m,i}=\frac{1}{3}\sum_{z=1}^{3}Y_{m,i,z},
\qquad
A_m=\frac{100}{500}\sum_{i=1}^{500}\bar{Y}_{m,i}.
\end{equation}
For each contrast, 10,000 paired problem-bootstrap resamples retain all
methods and all three generations for every sampled problem. The reported
95\% intervals therefore quantify evaluation variation across problems,
conditional on these training runs and sampled generations. They do not
measure training-seed uncertainty.

\begin{table}[htbp]
\centering
\small
\caption{Meta-Llama-3-8B on MATH-500 with matched learning rate
$5\times10^{-5}$. Accuracy and generation-cap hit rate average three
generation seeds; each method has one training seed.}
\label{tab:math_details}
\begin{tabular}{lrr}
\toprule
Method & Accuracy (\%) & Generation-cap hit (\%) \\
\midrule
FFT & 32.93 & 58.53 \\
MoLF & 31.53 & 60.07 \\
LoRA-128 & 20.60 & 70.20 \\
\bottomrule
\end{tabular}
\end{table}

\begin{table}[htbp]
\centering
\small
\caption{Paired MATH-500 differences and 95\% problem-bootstrap intervals,
in percentage points. These intervals are conditional on one training run
per configuration.}
\label{tab:math_contrasts}
\begin{tabular}{lrr}
\toprule
Contrast & Difference & 95\% interval \\
\midrule
MoLF $-$ FFT & $-1.40$ & $[-3.87,+1.00]$ \\
MoLF $-$ LoRA-128 & $+10.93$ & $[+8.40,+13.40]$ \\
\bottomrule
\end{tabular}
\end{table}

The MoLF--LoRA interval is above zero under this evaluation protocol.
The MoLF--FFT interval contains zero, which leaves its ordering uncertain
and does not establish equivalence. FFT and MoLF have similar cap-hit rates,
whereas LoRA reaches the cap more often. Thus, the MoLF--LoRA contrast can
reflect both mathematical-answer quality and termination within the allotted
generation window.\looseness=-1

A symmetric two-learning-rate training grid also includes $10^{-4}$.
At that rate, MoLF obtains 29.33\% and LoRA 21.87\% MATH-500 accuracy, with
cap-hit rates of 62.27\% and 68.47\%, respectively. FFT at $10^{-4}$ was
trained but not evaluated because the compute allocation was exhausted.
Consequently, the matched $5\times10^{-5}$ comparison is the complete
three-method result, rather than a comparison of independently optimized
methods. We do not use the unadapted model's score to estimate SFT gains:
the prompt and generation budget do not provide a reliable comparison
between the raw base checkpoint and the fine-tuned checkpoints.


\FloatBarrier
\subsection{Training efficiency: protocols and full measurements}
\label{app:efficiency}

Standard MoLF retains dense gradients and Adam moments alongside its LoRA pathway; Top-1 selection therefore does not provide LoRA-like training-memory savings. Table~\ref{tab:efficiency_context_summary} summarizes Qwen3-8B-Base on eight H100s at matched workloads within each context, with peak allocated memory reported per GPU. The protocols and complete measurements below distinguish this context sweep from a separate single-H100 Qwen2.5-7B benchmark.

\begin{table}[htbp]
\centering
\caption{Qwen3-8B-Base on eight H100 GPUs. Memory is peak PyTorch allocation per GPU; overheads compare methods at the same context and workload.}
\label{tab:efficiency_context_summary}
\begin{tabular}{rcccc}
\toprule
Context & FFT peak (GiB) & MoLF peak (GiB) & Memory overhead & Step-time overhead \\
\midrule
4,096  & 26.79 & 26.97 & $+0.7\%$  & $+19.6\%$ \\
8,192  & 28.05 & 28.23 & $+0.6\%$  & $+17.9\%$ \\
16,384 & 30.56 & 32.13 & $+5.1\%$  & $+14.2\%$ \\
32,768 & 36.97 & 40.89 & $+10.6\%$ & $+9.4\%$ \\
\bottomrule
\end{tabular}
\end{table}

\paragraph{Single-GPU 7B benchmark.}
We measure Qwen2.5-7B on one exclusive 80-GiB H100 using fixed CounterFact
batches. All four methods use BF16 parameters and Adam moments, microbatch
size 1, four-step accumulation, and sequence length 128, corresponding to
512 padded tokens per optimizer step. We reuse the same 160 pre-materialized
examples and use TF32, scaled dot-product attention, and gradient
checkpointing. There is no distributed training, offloading, quantization,
or compilation. LoRA uses one rank-128 RS-LoRA adapter per adapted module;
MoLF adds the same adapter to a trainable dense pathway; MoLF-E freezes the
dense model and uses two rank-128 adapters. Adapter scaling and dropout are
$32$ and $0.05$.

Each method is run in three fresh processes, arranged in three randomized
four-method blocks on the same GPU. Each run has 10 warmup optimizer steps
and 30 timed steps. Warmup materializes the optimizer states; synchronization
at the timing boundaries measures a complete accumulation-and-update window.
Evaluation, checkpointing, and per-step routing logs are disabled, while
routing computation remains enabled. Peak allocation and reservation are
recorded over both startup and steady-state windows; the larger is the
training peak. PyTorch allocator measurements exclude CUDA context and
allocations outside PyTorch.

\begin{table}[htbp]
\centering
\small
\caption{Qwen2.5-7B at 512 padded tokens per optimizer step on one H100.
Time is mean $\pm$ sample SD over three fresh processes. State and peak
memory are in GiB; trainable \mbox{parameter counts are in millions.}}
\label{tab:efficiency_7b_full}
\begin{tabular}{lrrrr}
\toprule
Method & Trainable params & Adam state & Peak alloc./resv. & Seconds/step \\
\midrule
FFT & 7,615.6 & 28.37 & 71.20 / 71.78 & $0.5837\pm0.0046$ \\
LoRA & 323.0 & 1.20 & 17.36 / 18.02 & $0.7161\pm0.0083$ \\
MoLF & 7,938.6 & 29.57 & 61.42 / 62.57 & $0.9562\pm0.0063$ \\
MoLF-E & 645.9 & 2.41 & 19.38 / 20.08 & $1.0294\pm0.0080$ \\
\bottomrule
\end{tabular}
\end{table}

MoLF adds approximately 323M trainable parameters and 1.20 GiB of optimizer
state to FFT and has 63.8\% greater step time at this operating point.
LoRA and MoLF-E have 22.7\% and 76.4\% greater step time, respectively.
MoLF's lower observed peak allocation is caused by implementation-dependent
transient allocations: FFT's foreach AdamW processes a large parameter
group, while MoLF processes smaller module-local groups. It is not a
persistent-state memory saving. The recorded optimizer state directly shows
the dense-state cost of standard MoLF. In contrast, LoRA and MoLF-E reduce
training memory by freezing the dense model.

\paragraph{Eight-GPU 8B context sweep.}
We benchmark Qwen3-8B-Base on one eight-H100 node at context lengths
4,096, 8,192, 16,384, and 32,768. The 12 method/context configurations run
sequentially on the same node. Microbatch size 1, accumulation 4, and world
size 8 give 32 sequences per optimizer step at every length; the token
count therefore scales with context. This stack uses FSDP2, FP32 master
weights, global gradient-norm clipping, and compilation. Each configuration
has 14 steps; the first four are discarded for compilation and allocator
warmup, and the remaining ten determine the step-time mean and coefficient
of variation (CV). Reported throughput is total padded tokens per second
across the eight devices, while memory is per device. Contexts beyond the
native 32,768-token window are not measured.

\begin{table}[htbp]
\centering
\small
\caption{Full Qwen3-8B-Base context sweep on eight H100s. Timing CV is
computed across the ten timed steps within the primary run. Peak allocated
and reserved memory are per GPU in GiB.}
\label{tab:efficiency_8b_full}
\begin{tabular}{rlrrrrr}
\toprule
Context & Method & s/step & CV (\%) & Tokens/s & Peak alloc. & Peak resv. \\
\midrule
4,096 & FFT & 3.196 & 2.08 & 41,013 & 26.79 & 38.56 \\
      & LoRA & 3.850 & 3.15 & 34,044 & 12.38 & 35.69 \\
      & MoLF & 3.821 & 2.49 & 34,304 & 26.97 & 40.04 \\
\midrule
8,192 & FFT & 5.663 & 0.37 & 46,292 & 28.05 & 38.56 \\
      & LoRA & 7.612 & 1.41 & 34,439 & 14.46 & 35.69 \\
      & MoLF & 6.678 & 0.67 & 39,252 & 28.23 & 40.04 \\
\midrule
16,384 & FFT & 12.374 & 0.11 & 42,371 & 30.56 & 38.56 \\
       & LoRA & 16.785 & 0.47 & 31,236 & 18.64 & 35.69 \\
       & MoLF & 14.134 & 0.21 & 37,093 & 32.13 & 40.04 \\
\midrule
32,768 & FFT & 32.621 & 0.05 & 32,144 & 36.97 & 43.44 \\
       & LoRA & 41.604 & 0.10 & 25,204 & 27.00 & 35.69 \\
       & MoLF & 35.695 & 0.16 & 29,376 & 40.89 & 47.42 \\
\bottomrule
\end{tabular}
\end{table}

\begin{table}[htbp]
\centering
\small
\caption{Relative costs in the Qwen3-8B-Base context sweep. Each comparison
uses FFT at the same context length as its denominator.}
\label{tab:efficiency_8b_relative}
\begin{tabular}{rrrr}
\toprule
Context & MoLF time overhead & LoRA time overhead & MoLF allocation overhead \\
\midrule
4,096 & $+19.6\%$ & $+20.5\%$ & $+0.7\%$ \\
8,192 & $+17.9\%$ & $+34.4\%$ & $+0.6\%$ \\
16,384 & $+14.2\%$ & $+35.6\%$ & $+5.1\%$ \\
32,768 & $+9.4\%$ & $+27.5\%$ & $+10.6\%$ \\
\bottomrule
\end{tabular}
\end{table}

The full sweep is repeated on the same node. Table~\ref{tab:efficiency_repeat}
reports relative changes in step time; peak allocation agrees to 0.01 GiB
for every configuration. The maximum within-run CV is 3.15\%, whereas the
largest between-run step-time deviation is 0.47\%. These quantify different
sources of measurement dispersion and are not confidence intervals.

\begin{table}[htbp]
\centering
\small
\caption{Step-time change in the same-node replicate relative to the
primary Qwen3-8B-Base sweep.}
\label{tab:efficiency_repeat}
\begin{tabular}{rrrr}
\toprule
Context & FFT & LoRA & MoLF \\
\midrule
4,096 & $+0.23\%$ & $-0.47\%$ & $+0.06\%$ \\
8,192 & $+0.47\%$ & $-0.37\%$ & $+0.21\%$ \\
16,384 & $+0.25\%$ & $-0.26\%$ & $+0.30\%$ \\
32,768 & $+0.03\%$ & $-0.05\%$ & $-0.04\%$ \\
\bottomrule
\end{tabular}
\end{table}

\paragraph{Interpretation and measurement boundaries.}
At a fixed context, all methods process the same number of tokens per step,
so the step-time ratios determine relative throughput. MoLF's overhead
relative to FFT decreases from 19.6\% to 9.4\% over this sweep, consistent
with fixed routing and moment-management costs being amortized as token
computation increases. Standard MoLF's peak allocation remains near or above
FFT. Reserved memory also reflects the caching allocator's behavior, so we
use allocated memory for the cross-method memory comparison.

The 7B benchmark and 8B sweep differ in model, token count, precision, and
distributed/software stack; neither yields a universal overhead number.
LoRA is implemented using PEFT, whereas MoLF uses its own wrapper and
optimizer, so timing also reflects implementation differences. MoLF's
observed speed relative to PEFT LoRA at longer contexts does not imply an
inherent speed ordering between dense and low-rank adaptation. The 4,096-token
MoLF--LoRA time difference is smaller than their within-run CVs. These are
training measurements, not comparisons of FLOPs, energy, or consumer-device
feasibility. MoLF-E is measured only in the single-GPU short-context benchmark,
where its lower memory use does not translate into faster training. These
measurements motivate systems optimization and distinguish standard MoLF's
update-selection purpose from MoLF-E's memory-saving purpose.

\FloatBarrier
\section{Routing Controls and Projection Specialization}
\label{app:routing_controls}

\subsection{Matched EPD and coin-flip Top-1 routing}
\label{app:matched_routing}

\paragraph{What the control isolates.}
Updating both pathways ($K=2$) changes both the selection rule and the number
of pathways physically updated. To separate EPD-based selection from the
Top-1 constraint, we compare EPD with an independent fair coin flip at every
module and optimizer step. Both arms run all pathways in forward/backward
computation and update both pathways' AdamW moments and counters. Exactly
one pathway per module receives the physical AdamW update, including
weight decay. The control matches this pathway-update count; it does not
force the same FFT/LoRA allocation ratio or the same number of scalar
parameters to change.

\paragraph{Protocol.}
We use Qwen2.5-3B on MedMCQA with five paired training seeds, 42--46. The 36
layers contain 252 wrapped projection modules, each with one dense and one
rank-128 RS-LoRA pathway. Within each seed, both arms start from the same
saved pre-optimizer initialization and use the same hashed sample-order
manifest. The coin-flip schedules are generated and hashed before training.
Each arm uses two epochs, effective batch size 32, and 11,428 optimizer
steps. Both pathway learning rates are $5\times10^{-5}$, with a linear
schedule and warmup ratio 0.1. AdamW uses $(0.9,0.999)$ and
$\epsilon=10^{-8}$, maximum gradient norm 1.0, dense/LoRA decay
$0.1/0.01$, and LoRA scaling/dropout $32/0.05$. We evaluate the full
4,183-example validation split. These matched fresh runs are distinct from
the original $K=2$ and PFN ablations.

\begin{table}[htbp]
\centering
\small
\caption{EPD versus independent coin-flip Top-1 routing on Qwen2.5-3B
MedMCQA. Every arm updates exactly one pathway per module and step. The
last row is a descriptive sensitivity analysis excluding seed 44, not an
alternative primary estimate. Differences are in percentage points.}
\label{tab:matched_routing_seeds}
\begin{tabular}{lrrr}
\toprule
Seed & EPD accuracy (\%) & Coin-flip accuracy (\%) & Difference \\
\midrule
42 & 61.1045 & 60.0287 & $+1.0758$ \\
43 & 61.2718 & 59.4071 & $+1.8647$ \\
44 & 60.1960 & 33.4449 & $+26.7511$ \\
45 & 59.7179 & 60.6024 & $-0.8845$ \\
46 & 60.3634 & 59.6940 & $+0.6694$ \\
\midrule
Mean $\pm$ SD & $60.5307\pm0.6479$ & $54.6354\pm11.8542$ & $+5.8953\pm11.7016$ \\
Without seed 44 & $60.6144\pm0.7163$ & $59.9331\pm0.5135$ & $+0.6813\pm1.1558$ \\
\bottomrule
\end{tabular}
\end{table}

EPD has higher accuracy in four of five seeds and a mean advantage of
$5.8953$ points. The coin-flip collapse at seed 44 contributes strongly:
without that seed, the mean difference is $0.6813$ points, with EPD higher
in three of four pairs. These results favor informed selection in this
setting while leaving its average magnitude uncertain; a single collapse
does not establish general stability. The experiment also does not separate
the effects of EPD's module-level allocation fractions from the timing of
individual choices, and it does not test a coordinate-sparse update pathway.

\FloatBarrier
\subsection{Projection-level selection statistics}
\label{app:projection_routing}

Table~\ref{tab:projection_routing} reports the range, over five seeds in
each reported task setting, of the fraction of physical updates assigned
to LoRA; the remaining updates go to FFT. These summaries describe
persistent projection-level specialization rather than a single global
FFT/LoRA choice.

\begin{table}[htbp]
\centering
\small
\caption{LoRA selection frequencies by projection family, reported as
minimum--maximum percentages across five EPD seeds per task setting.
These are descriptive summaries, not \mbox{a paired intervention on task.}}
\label{tab:projection_routing}
\begin{tabular}{lrrl}
\toprule
Projection & Med: LoRA updates & Fact: LoRA updates & Observed preference \\
\midrule
\texttt{k\_proj} & $0.00\%$ & $0.00\%$ & FFT throughout \\
\texttt{v\_proj} & $0.00\%$ & $0.00\%$ & FFT throughout \\
\texttt{down\_proj} & $5.20$--$7.65\%$ & $6.94$--$8.11\%$ & Mostly FFT \\
\texttt{o\_proj} & $61.97$--$69.23\%$ & $68.27$--$73.18\%$ & Mostly LoRA \\
\texttt{q\_proj} & $76.53$--$82.83\%$ & $95.57$--$96.51\%$ & LoRA, varying by setting \\
\texttt{gate\_proj} & $99.36$--$99.50\%$ & $99.80$--$99.87\%$ & Almost always LoRA \\
\texttt{up\_proj} & $99.59$--$99.76\%$ & $99.91$--$99.93\%$ & Almost always LoRA \\
\bottomrule
\end{tabular}
\end{table}

FFT is selected for every recorded key/value projection update, while LoRA
receives over 99\% of gate/up projection updates. Query projections show
different selection fractions in the two settings. These regularities
are consistent with module-local routing, but selection frequencies alone
do not show that each preference is optimal or caused by the task. EPD is
also sensitive to parameterization and gradient scaling, so structural
differences among projection families can contribute \mbox{to the observed pattern.}

\FloatBarrier
\subsection{Relation to the original controls and routing trajectories}
\label{app:routing_interpretation}

The original $K=2$ comparison in
Section~\ref{subsubsec:ablation_expert_selection} uses both physical updates
per module. Top-1 improves all three Med results while the SQL differences
are small and mixed, but this comparison jointly changes routing and update
count. The PFN comparison uses an author-constructed diagnostic based on
the root-mean-square preconditioned Adam direction
(Appendix~\ref{app:pfn_details}). EPD improves on or ties PFN in five of six
reported model--task cells; the largest gains are approximately 3.1 points on Gemma-3-1B
Med and 1.46 on Qwen2.5-3B Med. Gemma-3-1B SQL is tied, and Qwen2.5-1.5B
SQL favors PFN by 0.48 points. These original controls have no repeated-seed
uncertainty and are not substitutes for the fresh matched Top-1 comparison.

The routing trajectories in Appendix~\ref{subsec:router_behavior} provide
complementary descriptive evidence. Aggregate FFT/LoRA proportions often
settle within approximately 500 optimizer steps, while module heatmaps show
persistent assignments with a smaller set of switching modules. There is no
universal sequence such as LoRA early and FFT late. Med is LoRA-dominated
across the three original backbones, whereas Fact and SQL routing varies
with backbone. Because moments remain current for both pathways even when
one is not physically updated, these observations are consistent with
persistent differences in their measured update scores. They do not
establish a causal explanation for performance or validate early routing
as a predictor of the best final static method.

\FloatBarrier
\section{Additional Results}
\subsection{Behavior of the Router}
\label{subsec:router_behavior}
For each fine-tuning setup, we examine the behavior of the router. The router acts as the score-based selector inside the MoLF optimizer: for each MoLF-wrapped linear module, it scores candidate experts using the EPD score derived from the module's Adam moments. Only the fine-tuning expert with a higher EPD score receives a parameter update at each step. In these experiments, the two candidates per module are the original base weight (FFT) and a rank-$128$ LoRA expert.

\begin{figure}[htbp]
\centering
\includegraphics[width=\linewidth]{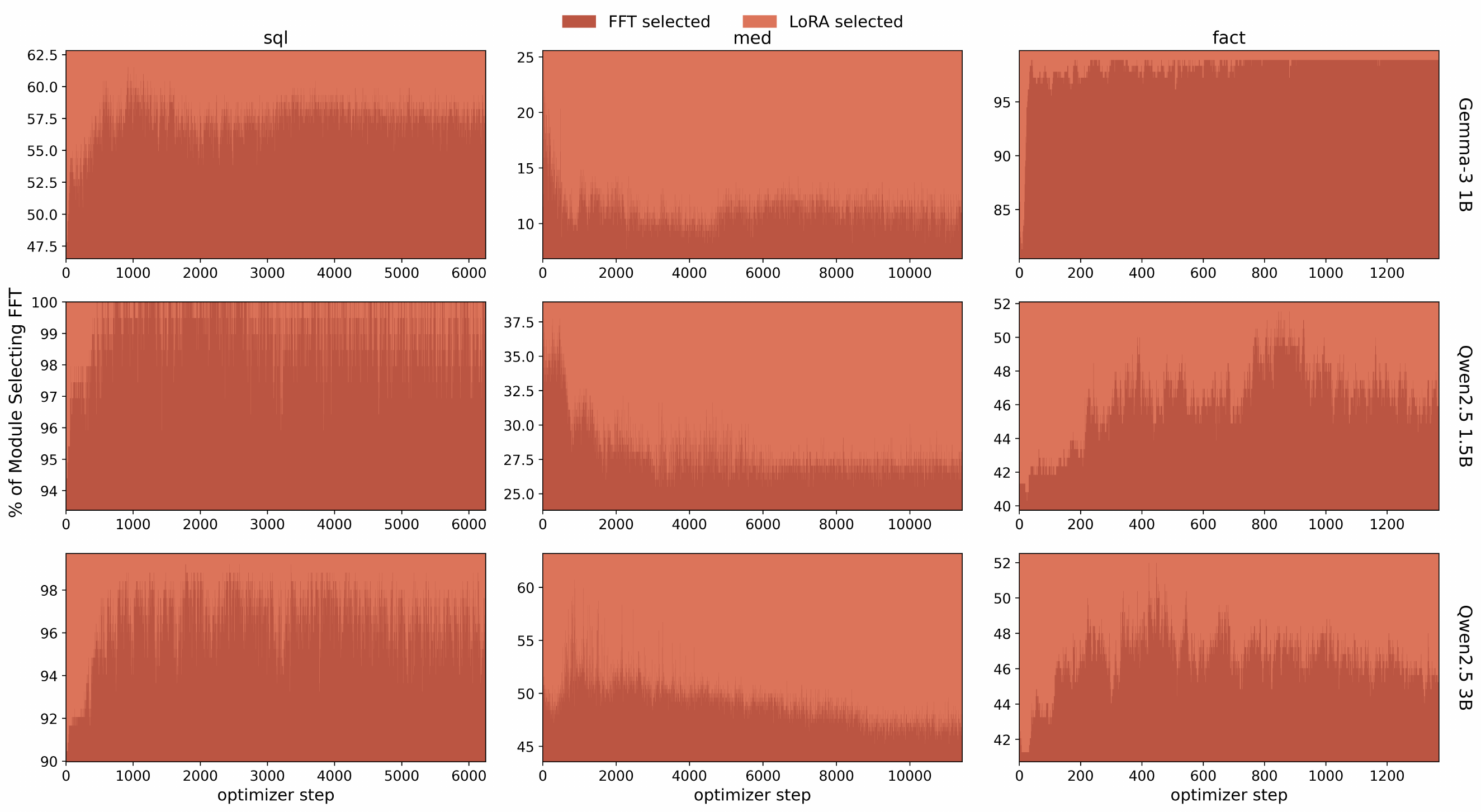}
\caption{Aggregate router decisions over training. Bars represent the percentage of modules selecting FFT (lower/dark) versus LoRA (upper/light) at each optimizer step. The y-axis for each panel is scaled to its operational range to highlight boundary fluctuations. The selected mixture varies with both the task and the model. Med is strongly LoRA-dominated across all scales, but the FFT preference on Fact and SQL inverts between architectures: on Gemma-3-1B, Fact leans heavily toward FFT while SQL is roughly balanced; on the Qwen2.5 models, SQL leans heavily toward FFT while Fact is roughly balanced. Despite this heterogeneity, the router converges to a stable global mixture early in training (typically within $500$ steps) and maintains it. We also observe that larger models (Qwen2.5-3B) exhibit a tighter, more compressed envelope of task-specific routing compared to smaller models (Gemma-3-1B).}
\label{fig:router_percent}
\end{figure}
\FloatBarrier
\begin{figure}[htbp]
\centering
\includegraphics[width=\linewidth]{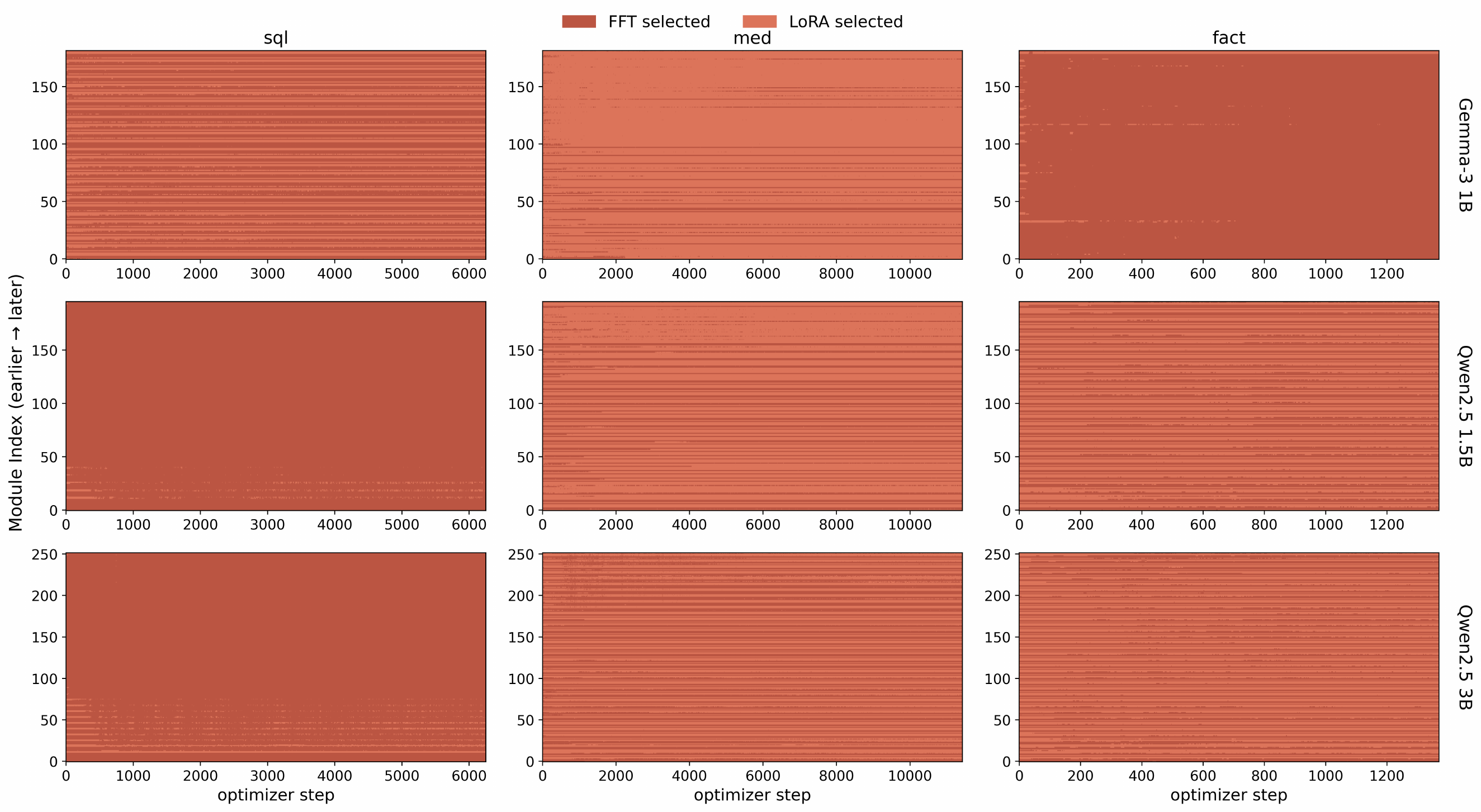}
\caption{Per-module router decisions over training. Rows index the modules in parameter order (bottom = earliest, top = latest), columns represent optimizer steps, and pixels encode the winning expert. The pronounced horizontal banding demonstrates that the aggregate mixtures in Figure~\ref{fig:router_percent} result from persistent per-module specialization, not from uniform step-wise alternation. The empirical row distribution is strongly bimodal: modules typically commit definitively to either FFT or LoRA early in the run. The minor step-wise fluctuations observed in the aggregate view are driven by a small population of swing modules, confirming that the router performs persistent structural assignments rather than continuous resampling.}
\label{fig:router_evolve}
\end{figure}

\subsection{MoLF-E Rank Configuration}
\label{subsec:ablation_rank}
\begin{figure}[htbp]
\centering
\includegraphics[width=0.9\textwidth]{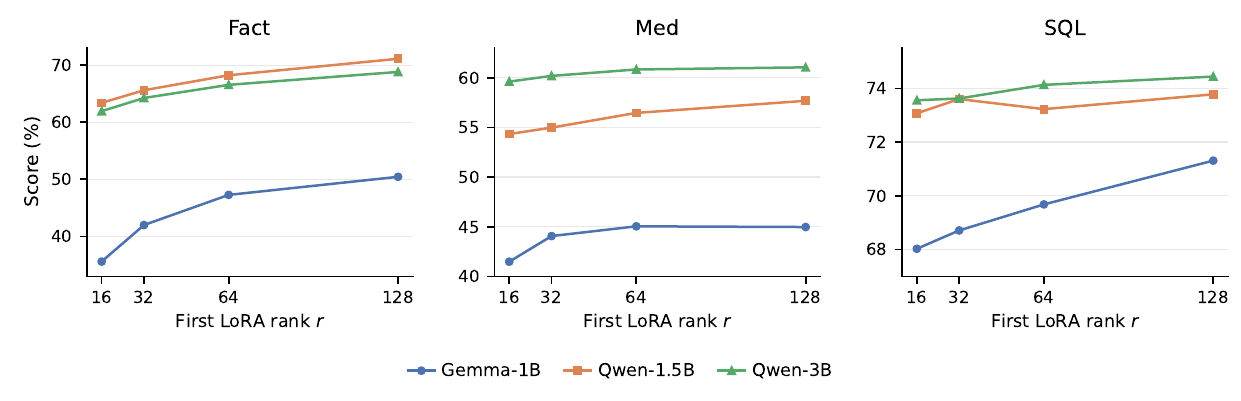}
\caption{Rank ablation of MoLF-E: performance versus the first LoRA expert's rank $r$, with the second expert fixed at rank $128$ and Top-$K{=}1$ routing. Each panel plots one task across all three models; the y-axis is Efficacy Score (\%) on Fact and accuracy (\%) on Med and SQL.}
\label{fig:ablation_rank}
\end{figure}

We also experimented with varying ranks of the LoRA experts in MoLF-E to isolate the effect of rank capacity on the three fine-tuning regimes. All configurations use two parallel LoRA experts with ranks $r$ and $128$, Top-$K{=}1$ routing, and MoLF-E's frozen base-weight setting. To match the parameter scope of standard PEFT baselines, we additionally freeze the non-linear parameters (embeddings, layer norms, and \texttt{lm\_head}), leaving only the LoRA experts trainable. Figure~\ref{fig:ablation_rank} reports the rank sweep.\looseness=-1

The sweep exhibits task-dependent rank sensitivity. On Fact, the Efficacy Score grows substantially with the first expert's rank $r$: by $14.85$ percentage points on Gemma-3-1B ($r=16 \to 128$), $7.73$ on Qwen2.5-1.5B, and $6.88$ on Qwen2.5-3B. Two effects are entangled with this change. First, the $r=16$ mixture underperforms a single static rank-16 LoRA on Gemma-3-1B Fact (Table~\ref{tab:fft_vs_lft}), so raw representational capacity alone is insufficient to explain the low-rank result. Second, under the simplifying gradient-scaling assumptions behind the approximate $\eta_t^{(i)}\alpha_i/\sqrt{r_i}$ dependence of EPD, the smaller expert has a $\sqrt{128/r}$ score advantage at equal learning rate and scaling constant. Capacity and routing preference therefore change together; the sweep does not establish that the preference is optimal. Rank-invariant scoring, rank-aware learning rates, and routing priors would help separate these effects, but have not been evaluated here. The fixed-rank FFT/LoRA PFN comparison does not supply this missing unequal-rank MoLF-E experiment.

On Med, accuracy increases mildly with rank and saturates near $r=64$ (Gemma-3-1B peaks at $45.04\%$ and then slightly declines). On SQL, the sweep is nearly flat on both Qwen2.5 models (less than one percentage point of spread) and increases mildly on Gemma-3-1B ($3.28$ points from $r=16$ to $128$). These are descriptive rank-sensitivity results. They do not show that Med's principal gradient components have been captured, establish capability preservation, or prove a concentrated optimal-update spectrum for SQL. The endpoint analysis in Appendix~\ref{app:spectral} explains why those stronger spectral interpretations require additional evidence.

\section{Licenses for Existing Assets}
\label{sec:licenses}

Our experiments use publicly released pre-trained language models, fine-tuning and evaluation datasets, and an answer-verification model.

Gemma-3-1B-PT is released under the Gemma Terms of Use (\url{https://ai.google.dev/gemma/terms}), which permits research and downstream fine-tuning subject to Google's prohibited use policy; we use the model only for academic fine-tuning experiments and do not redistribute model weights. Qwen2.5-1.5B is released under the Apache License 2.0 (\url{https://huggingface.co/Qwen/Qwen2.5-1.5B/blob/main/LICENSE}), which permits research and commercial use with attribution. Qwen2.5-3B is released under the Qwen Research License Agreement (\url{https://huggingface.co/Qwen/Qwen2.5-3B/blob/main/LICENSE}), which permits non-commercial research use with attribution; our use is strictly for academic research and does not constitute commercial deployment.

Meta-Llama-3-8B~\citep{grattafiori2024llama} is released under the Meta Llama 3 Community License Agreement (\url{https://huggingface.co/meta-llama/Meta-Llama-3-8B/blob/main/LICENSE}), subject to its terms and incorporated Acceptable Use Policy. Qwen2.5-7B~\citep{yang2024qwen2_5} is released under the Apache License 2.0 (\url{https://huggingface.co/Qwen/Qwen2.5-7B/blob/main/LICENSE}). Qwen3-8B-Base~\citep{qwen3technicalreport} is listed under the Apache License 2.0 in its official model card (\url{https://huggingface.co/Qwen/Qwen3-8B-Base}). We acknowledge the use of these pre-trained models with the notices required by their respective licenses.

For datasets, CounterFact is released as part of the ROME repository under the MIT License (\url{https://github.com/kmeng01/rome/blob/main/LICENSE}); MedMCQA is released under the Apache License 2.0 (\url{https://huggingface.co/datasets/openlifescienceai/medmcqa}); the Gretel Synthetic Text-to-SQL dataset is released by Gretel AI under the Apache License 2.0 (\url{https://huggingface.co/datasets/gretelai/synthetic_text_to_sql}). ODA-Mixture-100k~\citep{opendataarena2026mixture} is listed under the Apache License 2.0 in its official dataset card (\url{https://huggingface.co/datasets/OpenDataArena/ODA-Mixture-100k/blob/main/README.md}). It re-packages LIMO (Apache License 2.0; \url{https://huggingface.co/datasets/GAIR/LIMO}) and AM-Thinking-v1-Distilled (\url{https://huggingface.co/datasets/a-m-team/AM-Thinking-v1-Distilled}), whose dataset card specifies no license but restricts use to research purposes and prohibits commercial use.

We obtain MATH-500~\citep{huggingfaceh4math500} from HuggingFaceH4/MATH-500 (\url{https://huggingface.co/datasets/HuggingFaceH4/MATH-500/blob/main/README.md}), whose dataset card identifies its source as the 500-problem MATH evaluation split released with OpenAI's PRM800K~\citep{lightman2023lets}. The dataset card does not specify a license; the upstream PRM800K repository carries the MIT License (\url{https://github.com/openai/prm800k/blob/main/LICENSE}).

We judge MATH-500 answers with xVerify~\citep{chen2025xverify} (\url{https://github.com/IAAR-Shanghai/xVerify}), whose code and model checkpoints are released under CC BY-NC-ND 4.0; we use it only for non-commercial research evaluation and do not redistribute it.

We use these datasets only for non-commercial academic fine-tuning and evaluation, cite their sources in the paper, and do not redistribute them.

\section{Broader Impacts}
\label{sec:broader_impacts}
This research introduces MoLF and its memory-efficient variant, MoLF-E, to study optimizer-level selection among adaptation experts. MoLF may help practitioners navigate the choice between full-parameter and low-rank updates, but the original experiments still used hyperparameter selection and standard MoLF has measured runtime overhead and dense optimizer-state requirements. We therefore do not claim demonstrated reductions in total tuning cost, energy use, or carbon emissions. MoLF-E reduces memory by freezing dense weights, which can broaden the range of hardware on which the studied low-rank routing approach fits. These methods may facilitate adaptation of open-weight models to specialized domains, including medical question answering and structured data, subject to task-specific evaluation and resource constraints.

Conversely, because MoLF streamlines the fine-tuning process, it inadvertently lowers the barrier for malicious actors to specialize models for harmful purposes, such as generating biased content, disinformation, and spam. However, as an application-agnostic optimization method, MoLF shares these dual-use vulnerabilities with all fine-tuning advancements; thus, such risks are more appropriately addressed through rigorous base-model release policies, dataset curation, and strict access controls rather than by restricting algorithmic progress. Additionally, while the efficiency of MoLF-E enables decentralized, on-device training that enhances user privacy, it simultaneously complicates centralized auditing. Therefore, deploying these methods in sensitive, high-stakes environments necessitates comprehensive safety evaluations, strict monitoring, and sustained human oversight.


\newpage
\section*{NeurIPS Paper Checklist}

\begin{enumerate}

\item {\bf Claims}
    \item[] Question: Do the main claims made in the abstract and introduction accurately reflect the paper's contributions and scope?
    \item[] Answer: \answerYes{}
    \item[] Justification: We elucidate our contributions in the abstract and introduction, and provide comprehensive experiment results to support our claims.

\item {\bf Limitations}
    \item[] Question: Does the paper discuss the limitations of the work performed by the authors?
    \item[] Answer: \answerYes{}
    \item[] Justification:  We discuss the limitations in Section~\ref{sec:conclusion}.

\item {\bf Theory assumptions and proofs}
    \item[] Question: For each theoretical result, does the paper provide the full set of assumptions and a complete (and correct) proof?
    \item[] Answer: \answerNA{}
    \item[] Justification: Our study is mainly empirical exploration.

    \item {\bf Experimental result reproducibility}
    \item[] Question: Does the paper fully disclose all the information needed to reproduce the main experimental results of the paper to the extent that it affects the main claims and/or conclusions of the paper (regardless of whether the code and data are provided or not)?
    \item[] Answer: \answerYes{}
    \item[] Justification: We provide details in Section~\ref{subsec:setup} and Appendix~\ref{subsec:hyperparam_sweep}.

\item {\bf Open access to data and code}
    \item[] Question: Does the paper provide open access to the data and code, with sufficient instructions to faithfully reproduce the main experimental results, as described in supplemental material?
    \item[] Answer: \answerYes{}
    \item[] Justification: The datasets evaluated in our study are publicly available, as detailed in Section~\ref{subsec:setup}. Anonymized source code with a PyTorch implementation of our method is provided in the supplementary material.

\item {\bf Experimental setting/details}
    \item[] Question: Does the paper specify all the training and test details (e.g., data splits, hyperparameters, how they were chosen, type of optimizer) necessary to understand the results?
    \item[] Answer: \answerYes{}
    \item[] Justification: The training details can be found in Section~\ref{subsec:setup} and Appendix~\ref{subsec:hyperparam_sweep}.

\item {\bf Experiment statistical significance}
    \item[] Question: Does the paper report error bars suitably and correctly defined or other appropriate information about the statistical significance of the experiments?
    \item[] Answer: \answerNo{}
    \item[] Justification: Given the scale of the language models evaluated and the substantial computational cost associated with the fine-tuning regimes, conducting multiple independent training runs per configuration to establish confidence intervals was computationally prohibitive.

\item {\bf Experiments compute resources}
    \item[] Question: For each experiment, does the paper provide sufficient information on the computer resources (type of compute workers, memory, time of execution) needed to reproduce the experiments?
    \item[] Answer: \answerYes{}
    \item[] Justification: We list the GPU type used in Section~\ref{subsec:setup} and report per-run wall-clock time and total project compute (approximately 2500 H100 GPU-hours, including the hyperparameter sweep, ablations, and configurations not reported) in Appendix~\ref{subsec:compute}.
    
\item {\bf Code of ethics}
    \item[] Question: Does the research conducted in the paper conform, in every respect, with the NeurIPS Code of Ethics \url{https://neurips.cc/public/EthicsGuidelines}?
    \item[] Answer: \answerYes{}
    \item[] Justification: We have carefully reviewed our research, ensuring it conforms with the NeurIPS Code of Ethics.

\item {\bf Broader impacts}
    \item[] Question: Does the paper discuss both potential positive societal impacts and negative societal impacts of the work performed?
    \item[] Answer: \answerYes{}
    \item[] Justification: We discuss the broader societal impacts of our work in Appendix~\ref{sec:broader_impacts}.
    
\item {\bf Safeguards}
    \item[] Question: Does the paper describe safeguards that have been put in place for responsible release of data or models that have a high risk for misuse (e.g., pre-trained language models, image generators, or scraped datasets)?
    \item[] Answer: \answerNo{}
    \item[] Justification: The proposed method and empirical study, including the released code, is intended strictly for research purposes rather than direct industrial application.

\item {\bf Licenses for existing assets}
    \item[] Question: Are the creators or original owners of assets (e.g., code, data, models), used in the paper, properly credited and are the license and terms of use explicitly mentioned and properly respected?
    \item[] Answer: \answerYes{}
    \item[] Justification: All pre-trained models and datasets used in our experiments are publicly available and properly cited in the main text. We discuss their licenses and our usage in Appendix~\ref{sec:licenses}.

\item {\bf New assets}
    \item[] Question: Are new assets introduced in the paper well documented and is the documentation provided alongside the assets?
    \item[] Answer: \answerYes{}
    \item[] Justification: Related codes will be released with suitable license.

\item {\bf Crowdsourcing and research with human subjects}
    \item[] Question: For crowdsourcing experiments and research with human subjects, does the paper include the full text of instructions given to participants and screenshots, if applicable, as well as details about compensation (if any)? 
    \item[] Answer: \answerNA{}
    \item[] Justification: Neither crowdsourcing nor human subject research is involved in this work.

\item {\bf Institutional review board (IRB) approvals or equivalent for research with human subjects}
    \item[] Question: Does the paper describe potential risks incurred by study participants, whether such risks were disclosed to the subjects, and whether Institutional Review Board (IRB) approvals (or an equivalent approval/review based on the requirements of your country or institution) were obtained?
    \item[] Answer: \answerNA{}
    \item[] Justification: Neither crowdsourcing nor human subject research is involved in this work.

\item {\bf Declaration of LLM usage}
    \item[] Question: Does the paper describe the usage of LLMs if it is an important, original, or non-standard component of the core methods in this research? Note that if the LLM is used only for writing, editing, or formatting purposes and does \emph{not} impact the core methodology, scientific rigor, or originality of the research, declaration is not required.
    \item[] Answer: \answerNo{}
    \item[] Justification: LLMs are only used for writing.

\end{enumerate}

\end{document}